\documentclass{egpubl}
\usepackage{eurova2026}

\WsPaper           % uncomment for final version of EG Workshop contribution
\usepackage[T1]{fontenc}
\usepackage{dfadobe}  

\usepackage{cite}  % comment out for biblatex with backend=biber
\BibtexOrBiblatex
\electronicVersion
\PrintedOrElectronic
\ifpdf \usepackage[pdftex]{graphicx} \pdfcompresslevel=9
\else \usepackage[dvips]{graphicx} \fi

\usepackage{egweblnk} 
\usepackage{amsfonts}
\usepackage{amsmath}
\usepackage{amssymb}
\usepackage{booktabs}
\usepackage{enumitem}
\usepackage{float}
\usepackage{mathtools}
\usepackage{microtype}
\usepackage{placeins}
\usepackage{wrapfig}
\usepackage{xcolor}

\newcommand{\papertitle}{Local Neighborhood Instability in Parametric Projections: \\ Quantitative and Visual Analysis}
\newcommand{\papertitleshort}{Local Neighborhood Instability in Parametric Projections: Quantitative and Visual Analysis}

\newcommand{\authorlist}{%
  Frederik L. Dennig$^1$\orcid{0000-0003-1116-8450} and
  Daniel A. Keim$^1$\orcid{0000-0001-7966-9740}%
}

\newcommand{\authorlistshort}{Dennig and Keim}

\newcommand{\affiliationlist}{$^1$University of Konstanz, Germany}

\newcommand{\osflink}{https://osf.io/t7uc3}

\newcommand{\githublink}{https://github.com/fredooo/paramproj-instab}

\AtBeginDocument{

}

\newcommand{\headbf}[1]{
  \noindent\textbf{#1}%
}

\newcommand{\headit}[1]{
  \smallskip
  \noindent\emph{#1}%
}

\newcommand{\compresslist}{
  \setlength{\itemsep}{1pt}
  \setlength{\parskip}{0pt}
  \setlength{\parsep}{0pt}	
}

\DeclareMathAlphabet{\mathpzc}{OT1}{pzc}{m}{it}

\newcommand{\verticalLabel}[2]{\rotatebox{90}{\parbox{#1}{\centering\textbf{#2}}}}

\newcommand{\horizontalLabel}[2]{\parbox{#1}{\centering\textbf{#2}}}

\DeclareMathOperator{\clip}{clip}

\title[\papertitleshort]%
      {\papertitle}

\author[\authorlistshort]
{\parbox{\textwidth}{\centering
        \authorlist
        }
        \\
{\parbox{\textwidth}{\centering
       \affiliationlist
       }
}
}
\begin{document}

\teaser{
\includegraphics[width=\linewidth]{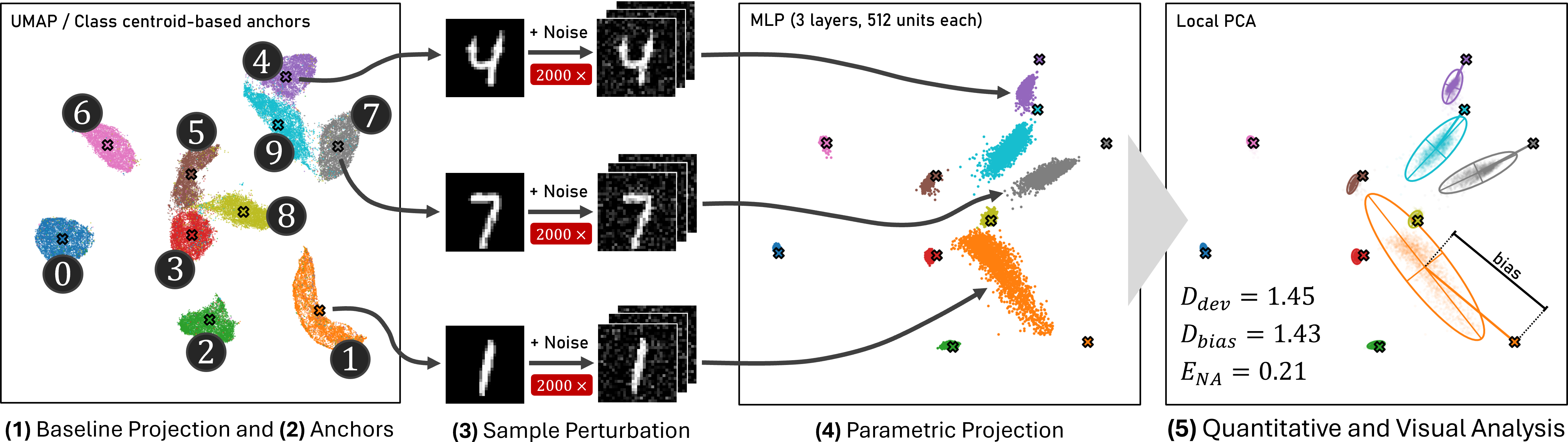}\\[-1em]
\caption{Framework overview on MNIST. \textbf{(1)}~A baseline UMAP projection is computed. \textbf{(2)}~Class centroid-based anchors are selected. \textbf{(3)}~Isotropic Gaussian noise is added to perturb anchor samples. \textbf{(4)}~A multi-layer perceptron (MLP) projects all noisy samples into 2D. \textbf{(5)}~Stability is assessed with three metrics: Mean displacement ($D_{\mathrm{dev}}$), displacement bias ($D_{\mathrm{bias}}$), and nearest-anchor assignment error ($E_{\mathrm{NA}}$); lower is better. Here, we show a local PCA visualization of displacement bias and the amplification or attenuation along principal directions.}
\label{fig:teaser}
}

\maketitle
%-------------------------------------------------------------------------
\begin{abstract}
%!TEX root = main.tex
Parametric projections let analysts embed new points in real time, but input variations from measurement noise or data drift can produce unpredictable shifts in the 2D layout. Whether and where a projection is locally stable remains largely unexamined. In this paper, we present a stability evaluation framework that probes parametric projections with Gaussian perturbations around selected anchor points and assesses how neighborhoods deform in the 2D embedding. Our approach combines quantitative measures of mean displacement, bias, and nearest-anchor assignment error with per-anchor visualizations of displacement vectors, local PCA ellipsoids, and Voronoi misassignment for detailed inspection. We demonstrate the framework's effectiveness on UMAP- and t-SNE-based neural projectors of varying network sizes and study the effect of Jacobian regularization as a gradient-based robustness strategy. We apply our framework to the MNIST and Fashion-MNIST datasets. The results show that our framework identifies unstable projection regions invisible to reconstruction error or neighborhood-preservation metrics.
%-------------------------------------------------------------------------
%  ACM CCS 1998
%  (see https://www.acm.org/publications/computing-classification-system/1998)
% \begin{classification} % according to https://www.acm.org/publications/computing-classification-system/1998
% \CCScat{Computer Graphics}{I.3.3}{Picture/Image Generation}{Line and curve generation}
% \end{classification}
%-------------------------------------------------------------------------
%  ACM CCS 2012
%   (see https://www.acm.org/publications/class-2012)
%The tool at \url{http://dl.acm.org/ccs.cfm} can be used to generate
% CCS codes.


\end{abstract}  
%-------------------------------------------------------------------------

\section{Introduction}\label{sec:introduction}

Making sense of high-dimensional data requires reducing it to a form humans can interpret.
Dimensionality reduction (DR) methods address this by mapping data to two or three dimensions
while preserving relationships such as distances and neighborhood structures~\cite{Nonato2019}.
A common shortcoming of these approaches is that they cannot represent the projection
as an explicit parametric function, preventing consistent and efficient projection
of new or synthetic data without recomputing the entire embedding~\cite{Maaten2009}.
In recent years, neural networks (NNs) have increasingly been used to learn such parametric
mappings~\cite{Espadoto2020Deep}.
These approaches can approximate non-linear projections and can generalize to a variety of DR
techniques~\cite{Dennig2025Evaluating}.
%Parametric projection methods address this limitation by learning an explicit mapping from the high-dimensional input space to a low-dimensional projection space.
%
With NNs, once trained, the mapping can be evaluated in constant time per point, enabling
fast out-of-sample projection of new or synthetic data without recomputing global pairwise
relationships~\cite{Sainburg2021}.
In contrast, classical DR techniques such as \emph{UMAP}~\cite{McInnes2018} and \emph{t-SNE}~\cite{Maaten2008Tsne} are non-parametric and require full recomputation when new data arrives.
This makes feed-forward multi-layer perceptron (MLP) projections attractive for \emph{streaming} and \emph{interactive settings}, for deployment
scenarios where embeddings must be produced on demand, and for workflows that require a consistent
layout across updates.
For visual analytics, this \emph{consistency} is essential, since analysts must be able to \emph{trust} that spatial patterns reflect data structures rather than artifacts of recomputation~\cite{Ngo2022Machine}.
However, parametric projections introduce a concern: The learned mapping is constrained only
by the training data and objective, and its behavior under small input perturbations is rarely examined~\cite{Dennig2025Evaluating}.
Current evaluation methods emphasize accurate local neighborhood preservation~\cite{Venna2001}, but ignore sensitivity to inputs that are slightly perturbed.
In this work, we analyze MLP projection architectures and how well they preserve local neighborhoods under input variations (see~\autoref{fig:teaser}).
Our evaluation combines \emph{quantitative measures} with \emph{visual analysis} to reveal systematic and sometimes surprising differences in how MLP-based parametric projections respond to small Gaussian perturbations.
In this paper, we contribute:
\begin{enumerate}[label=(\arabic*),left=0pt]
\compresslist
\item \emph{Three quantitative measures} to assess \emph{local neighborhood stability} in parametric projections under small input perturbations.
\item \emph{Three visualizations} that link our local stability measures to spatial and neighborhood changes in the projection.
\item \emph{Empirical comparisons} of parametric projections for UMAP and t-SNE, validating the effectiveness of stability assessment.
\item We share the analysis, results, and source code on \href{\osflink}{OSF} and \href{\githublink}{GitHub} for \emph{reproducibility}.
\end{enumerate}

\vspace{-1em}
\section{Related Work}\label{sec:related-work}

\headbf{Projection Methods for Visualization:}
We define a high-dimensional dataset as $D = \{x_i\}_{1 \leq i \leq n}$ with
$n$ samples $x_i \in \mathbb{R}^d$ and
a \emph{projection} method $P$ that maps $D$ to a lower-dimensional representation as
$P(D) = \{P(x_i) \mid x_i \in D\} = \{y_i\}_{1 \leq i \leq n}$, where
$P(D) \subset \mathbb{R}^q$ with $q \ll d$.
In our setting, $q = 2$, allowing $P(D)$ to be visualized as a two-dimensional
scatterplot.
Projection methods have been extensively studied and evaluated in several surveys
\cite{Cunningham2015,Espadoto2019,Nonato2019}. %Yin2007
They are commonly categorized as either \emph{linear} or \emph{non-linear}
methods~\cite{Shusen2016,Espadoto2019}, and further distinguished by whether they
primarily preserve \emph{global} structure or \emph{local} neighborhood relations.
Linear projection techniques such as PCA can be computed very
efficiently and are known to preserve global variance structure, while MDS
\cite{Kruskal1978Multidimensional} provides a non-linear alternative with a similar
global emphasis. %\cite{Jolliffe1986}
In contrast, many non-linear methods prioritize the preservation of local
neighborhoods at the expense of global structure, including t-SNE
\cite{Maaten2008Tsne} and UMAP~\cite{McInnes2018}.
%Autoencoder-based approaches are typically placed in this category of non-linear methods~\cite{wang2016auto}.
Among other projection quality measures~\cite{Espadoto2019}, Venna and Kaski~\cite{Venna2001} proposed \emph{Trustworthiness} and \emph{Continuity}, measuring intrusions and extrusions in local neighborhoods.

\vspace{-0.25em}
\headbf{Parametric Projections:} Standard non-linear DR methods~\cite{Kruskal1978Multidimensional,Maaten2008Tsne,McInnes2018} are non-parametric and typically require recomputation when projecting new data points~\cite{Hinterreiter2023}.
Parametric approaches address this limitation by learning explicit mapping functions, often implemented using neural networks (NNs), from the input space to a lower-dimensional embedding~\cite{Bunte2012}.
Van der Maaten~\cite{Maaten2009} introduced \emph{parametric t-SNE} using a feed-forward NN to approximate the non-parametric embedding.
Similarly, \emph{parametric UMAP}~\cite{Sainburg2021} employs NNs, including autoencoder-based architectures, to replace the non-parametric optimization step.
By training NNs to directly infer low-dimensional coordinates, Espadoto et al.~\cite{Espadoto2020Deep} demonstrated that sufficiently expressive models can approximate a wide range of existing non-parametric DR techniques.
\emph{HyperNP}~\cite{Appleby2022} extends this idea by learning parametric approximations across DR hyperparameters (e.g., perplexity in t-SNE), enabling interactive exploration.
Finally, \emph{ParaDime}~\cite{Hinterreiter2023} formalizes the design of NN-based DR methods through a grammar for parametric DR.
Recent approaches use autoencoders to learn parametric mappings~\cite{Dennig2025Evaluating}.

\vspace{-0.25em}
\headbf{Robustness and Adversarial Perturbations:}
Adversarial robustness concerns carefully crafted input perturbations designed to cause maximal change in model output under minimal input changes (i.e., small-norm constraint)~\cite{Goodfellow2015Explaining}.
Recent work has highlighted that parametric DR models can be vulnerable to attacks~\cite{Fujiwara2025}.
Adversarial robustness is a training objective~\cite{Madry2018}.
Cohen et al.~\cite{Cohen2019Certified} show that classifiers robust to Gaussian input noise are certifiably robust to adversarial perturbations.
Fast adversarial training~\cite{Wong2020Fast} provides a baseline defense mechanism, while Kabaha and Drachsler-Cohen~\cite{Kabaha2024} introduced verification methods for computing global robustness bounds of neural networks.
Lin and Fukuyama~\cite{Lin2024} developed a framework for calibrating DR hyperparameters in the presence of noise, addressing overfitting issues in t-SNE and UMAP.
Beyond adversarial training, explicit architectural constraints can enforce smoothness.
Jacobian regularization~\cite{Jakubovitz2018} penalizes the Frobenius norm of the output Jacobian with respect to inputs.
Spectral normalization~\cite{Miyato2018Spectral} bounds network Lipschitz constants by constraining the spectral norm of weight matrices.

\vspace{-0.25em}
\headbf{Visual Analysis of Point Distributions:}
DR methods produce 2D scatterplots with visualization challenges, specifically in the context of representing clusters and overplotting.
Instead of rendering individual points, density plots color-encode aggregated point counts, typically smoothed via kernel density estimation~\cite{Silverman1986Density,Scott1992Multivariate}.
%Contour lines at fixed probability mass levels offer an interpretable alternative~\cite{Scott1992Multivariate}.
Ellis and Dix~\cite{Ellis2007ClutterReduction} provide a taxonomy of clutter reduction techniques including filtering, sampling, opacity adjustment, and spatial distortion.
Von Landesberger et al.~\cite{vonLandesberger2009TimeDependentPointClouds} summarize point distributions using geometric primitives including convex hulls and minimum spanning trees.
%Mayorga and Gleicher~\cite{Mayorga2013Splatterplots} introduced \emph{Splatterplots}, which combine per-class density splatting with explicit boundary rendering to distinguish overlapping groups.
Chen et al.~\cite{Chen2014ScatterplotSamplingDesignSpace} explored hierarchical multi-scale sampling that maintains relative densities and outliers.
%, while Hu et al.~\cite{Hu2019DataSamplingScatterplots} cast multi-class sampling as a set cover optimization problem.
Introducing visual abstractions, Liao et al.~\cite{Liao2018ClusterBasedScatterplots} replace dense point groups with aggregate visual marks such as ellipses or convex hulls, reducing clutter while conveying cluster shape.

\vspace{-1em}
\section{Stability Evaluation Framework}\label{sec:method}

Our framework evaluates local stability by measuring how a given MLP projection responds to small input perturbations around representative \emph{anchor points}, essentially answering the question: \emph{How stable is it under plausible measurement noise?}
Given a projection $f : \mathbb{R}^d \rightarrow \mathbb{R}^q$, a high-dimensional dataset, and a strategy to select anchor points:
(1)~We fit a baseline projection method (e.g., UMAP or t-SNE) to obtain embedded coordinates.
(2)~We choose a set of anchor points based on the chosen strategy.
When class labels are available (as in our datasets), selecting samples near class centroids in projection space ensures representative anchors rather than outliers.
(3)~We generate perturbed inputs around each anchor by adding isotropic Gaussian noise.
(4)~We project the perturbed samples and compute quantitative stability metrics.
(5)~We visualize the resulting point clouds to reveal local geometric structure.
This framework applies uniformly to MLP-based parametric projections.

\begin{figure}[t]
\includegraphics[width=\linewidth]{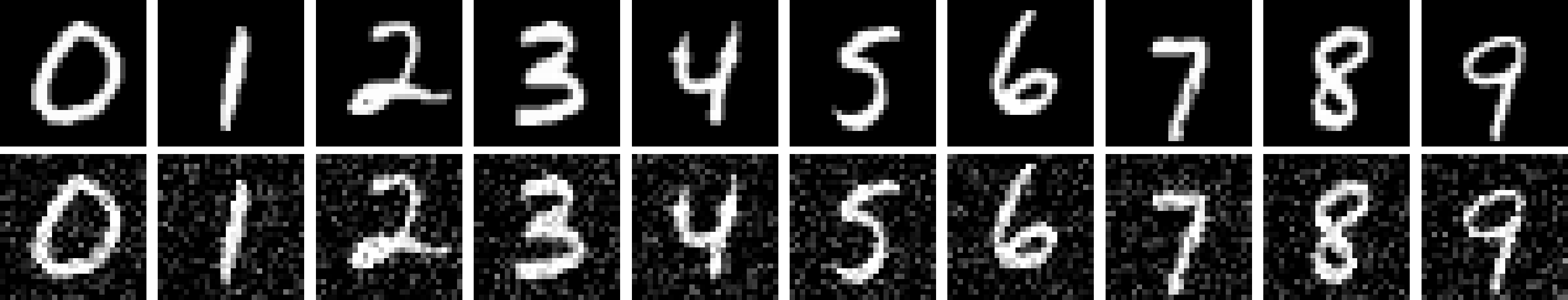}
\vspace{-1.5em}
\caption{Top -- Class centroid-based anchors of MNIST. Bottom -- Same images with isotropic Gaussian noise added ($\sigma=0.17$).}
\vspace{-2.0em}
\label{fig:noisy-mnist}
\end{figure}

\vspace{-0.25em}
\headbf{Local Neighborhood Stability:}
For each anchor point $x_0$, we probe local stability by adding isotropic Gaussian noise parameterized by a target input-space radius $r>0$.
Concretely, let $\varepsilon \sim \mathcal{N}(0,\sigma^2 I_d)$ in $d$ dimensions; then
\begin{equation}
r \coloneqq \sqrt{\mathbb{E}\!\left[\|\varepsilon\|_2^2\right]} = \sigma\sqrt{d}, \text{ with }
\sigma = \frac{r}{\sqrt{d}}, \text{ and }
\mathbb{E}\!\left[\|\varepsilon\|_2^2\right] = d\sigma^2,
\end{equation}
so that specifying a radius $r$ uniquely determines the noise scale.
In practice, $r$ is chosen to represent a small, interpretable neighborhood around each data point (e.g., such that perturbed samples remain visually
recognizable for image data).
As a heuristic, $r$ may be selected as a small fraction of a typical input-space distance (i.e., a very low percentile of all pairwise distances).
We choose the $0.25$th percentile.
For image data bounded to $[0,1]$, we clip perturbed values element-wise to this range (\autoref{fig:noisy-mnist}).
We generate $N$ perturbed samples per anchor and clip them to obtain $\tilde{x}_i = \clip(x_0 + \varepsilon_i)$.
Clipping reduces the realized perturbation magnitude relative to the nominal radius
$r$.
We report the effective (root-mean-square) radius
$r_{\mathrm{eff}} \coloneqq \sqrt{\frac{1}{N}\sum_{i=1}^{N}\|\tilde{x}_i - x_0\|_2^2}$,
computed from the clipped perturbations, the target radius $r$, and the resulting $\sigma$ in \autoref{tab:effective-r}.
The ratio $r_{\mathrm{eff}} / r$ quantifies the clipping effect.
Values near 1 indicate negligible clipping, while lower ratios reflect boundary saturation.

%\vspace{-0.15em}
\begin{table}[h]
\small
\centering
\begin{tabular}{lcccccc}
\textbf{Dataset} & $\mathbf{n}$ & $\mathbf{d}$ & $\mathbf{r}$ & $\mathbf{r}_{\mathrm{eff}}$ & $\mathbf{r_{\mathrm{eff}} / r}$ & $\sigma$ \\
\hline
\emph{MNIST}~\cite{Lecun1998} & 70000 & 784 & 4.73 & 3.46 & 0.73 & 0.17 \\
\emph{Fashion}~\cite{Xiao2017} & 70000 & 784 & 4.47 & 3.69 & 0.83 & 0.16 \\
\end{tabular}
\caption{
Size $\mathbf{n}$,
dimensionality $\mathbf{d}$,
noise radius $\mathbf{r}$,
effective noise radius $\mathbf{r_{\mathrm{eff}}}$,
the ratio $\mathbf{r_{\mathrm{eff}} / r}$, std. dev. $\sigma$. Noisy samples are created with the 0.25th pairwise-distance percentile, 5 anchors/class, and N=1000 per anchor, using more samples for $\mathbf{r_{\mathrm{eff}}}$ precision (\autoref{sec:evaluation}).
}
\label{tab:effective-r}
\vspace{-0.7em}
\end{table}

\vspace{-0.35em}
\headbf{Stability Measures:}
A stable projection maps similar inputs to similar outputs. Small perturbations in input space should yield small displacements in projection space, while unstable projections exhibit large, erratic, or systematically biased responses to minor input changes.
%behavior that undermines the reliability of visual analysis.
Thus, we define a clean anchor point $x_0$ with projected location $z_0 = f(x_0)$, and the projected perturbed samples $z_i = f(\tilde{x}_i)$ for $i = 1,\dots,N$. We propose three quantitative measures:

\vspace{-0.35em}
\headit{(1) Mean Displacement:}
The typical noise-induced drift at noise level $\sigma$ is quantified as
\begin{equation}
D_{\mathrm{dev}}(\sigma)
\coloneqq
\frac{1}{N}
\sum_{i=1}^{N}
\left\| z_i - z_0 \right\|_2 \approx \mathbb{E}[\left\| z - z_0 \right\|],
\end{equation}
where $N$ is the number of noisy samples.

\vspace{-0.25em}
\headit{(2) Displacement Bias:}
Systematic displacement of the mean projection from the anchor is measured as
\begin{equation}
D_{\mathrm{bias}}(\sigma)
\coloneqq
\left\|
\frac{1}{N}\sum_{i=1}^N z_i - z_0
\right\|_2 \approx
\left\| \mathbb{E}[z] - z_0 \right\|
\end{equation}

\vspace{-1em}
\headit{(3) Nearest-Anchor Assignment Error:}
For each anchor $z_0^{(a)}$ and its corresponding noise-perturbed projections $\{z_i^{(a)}\}_{i=1}^N$, we assign each projected point to its nearest anchor in the projected space:
\begin{equation}
\hat{a}(z) = \arg\min_{k} \, \| z - z_0^{(k)} \|_2 .
\end{equation}
We then define the assignment error at noise level $\sigma$ as
\begin{equation}
E_{\mathrm{NA}}(\sigma) \coloneqq \frac{1}{A} \sum_{a=1}^{A} \frac{1}{N} \sum_{i=1}^{N} \mathbf{1}\!\left[ \hat{a}\!\left(z_i^{(a)}\right) \neq a \right],
\end{equation}
where $A$ denotes the number of anchors and $\mathbf{1}[\cdot]$ is the indicator function.
This measure quantifies the probability that noise-induced perturbations cause a
sample to leave the Voronoi region of its original anchor, providing a direct notion of projection robustness.

Lower values of $D_{\mathrm{dev}}$, $D_{\mathrm{bias}}$, and $E_{\mathrm{NA}}$ indicate greater stability.
$D_{\mathrm{dev}}$ measures typical displacement magnitude; $D_{\mathrm{bias}}$ detects systematic drift in a consistent direction.
In our experiments, we use $N = 2000$ samples for computing $D_{\mathrm{dev}}$, $D_{\mathrm{bias}}$, and $E_{\mathrm{NA}}$.

\headbf{Visual Diagnostics:}
We propose three visualizations to assess the local neighborhood stability of parametric projections. In all visualizations, anchor points are visualized as crosses.

\vspace{-0.35em}
\begin{wrapfigure}[12]{l}{3.8cm} % lines, orientation, size
    \vspace*{-3.5mm} % translation for margins
    %\hspace*{-4.5mm}
    \setlength{\fboxsep}{0pt}
    \setlength{\fboxrule}{0.6pt}
    \fbox{\includegraphics[width=4.3cm]{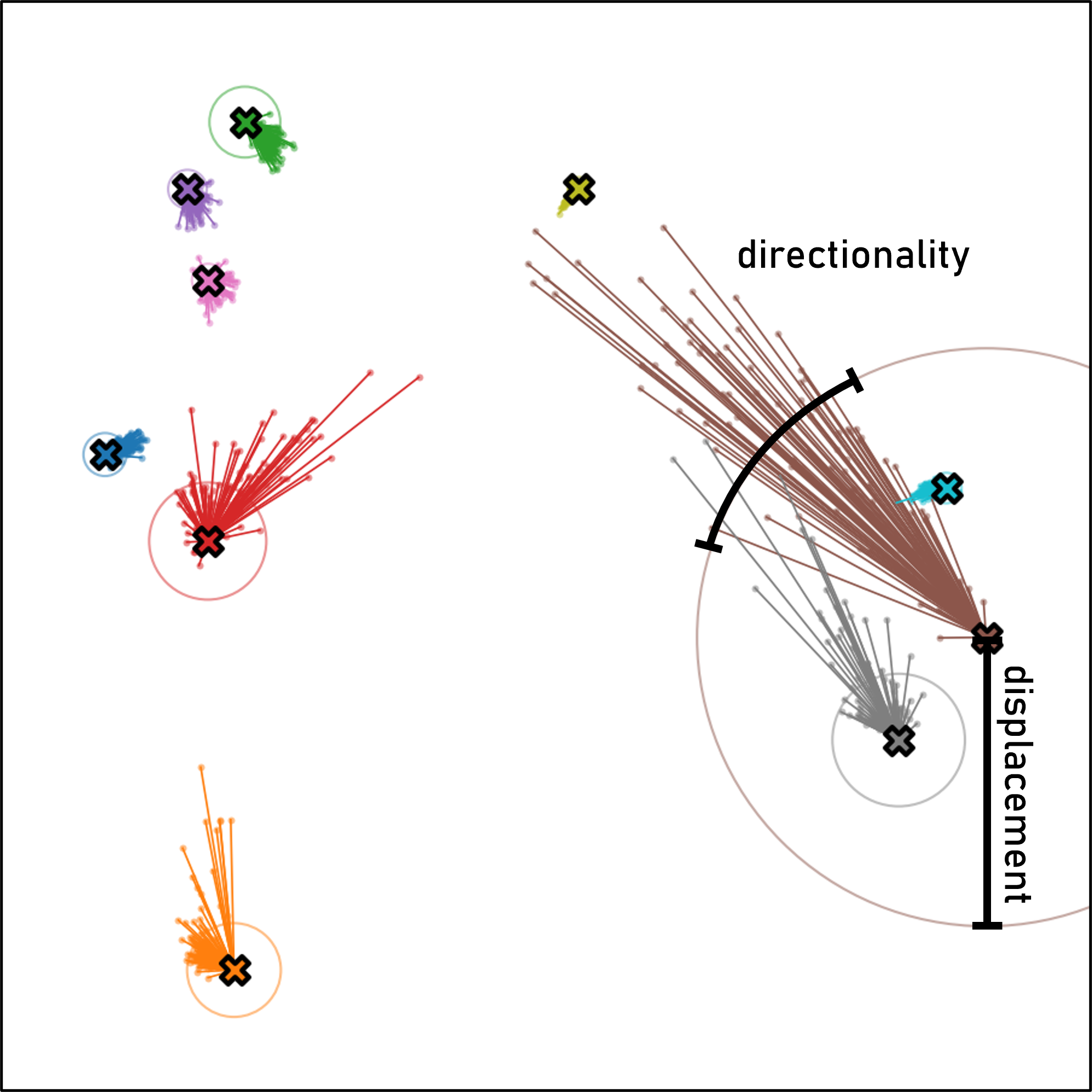}}
\end{wrapfigure}
\headit{Anchor Lines:} Anchor lines are geometric primitives~\cite{vonLandesberger2009TimeDependentPointClouds} and explicitly encode the \emph{relational} structure between each noisy projection and its corresponding anchor. By visualizing displacement vectors from the anchor to the noisy samples, this representation highlights directional bias, coherence, and asymmetry in the projected perturbations. It enables assessment of whether noise induces systematic drift away from the anchor or remains approximately isotropic. The \emph{mean displacement} ($D_{\mathrm{dev}}$) of the anchor is represented as a circle around the anchor point and captures the overall instability.

\vspace{-0.35em}
\begin{wrapfigure}[12]{l}{3.8cm} % lines, orientation, size
    \vspace*{-3.5mm} % translation for margins
    %\hspace*{-4.5mm}
    \setlength{\fboxsep}{0pt}
    \setlength{\fboxrule}{0.6pt}
    \fbox{\includegraphics[width=4.3cm]{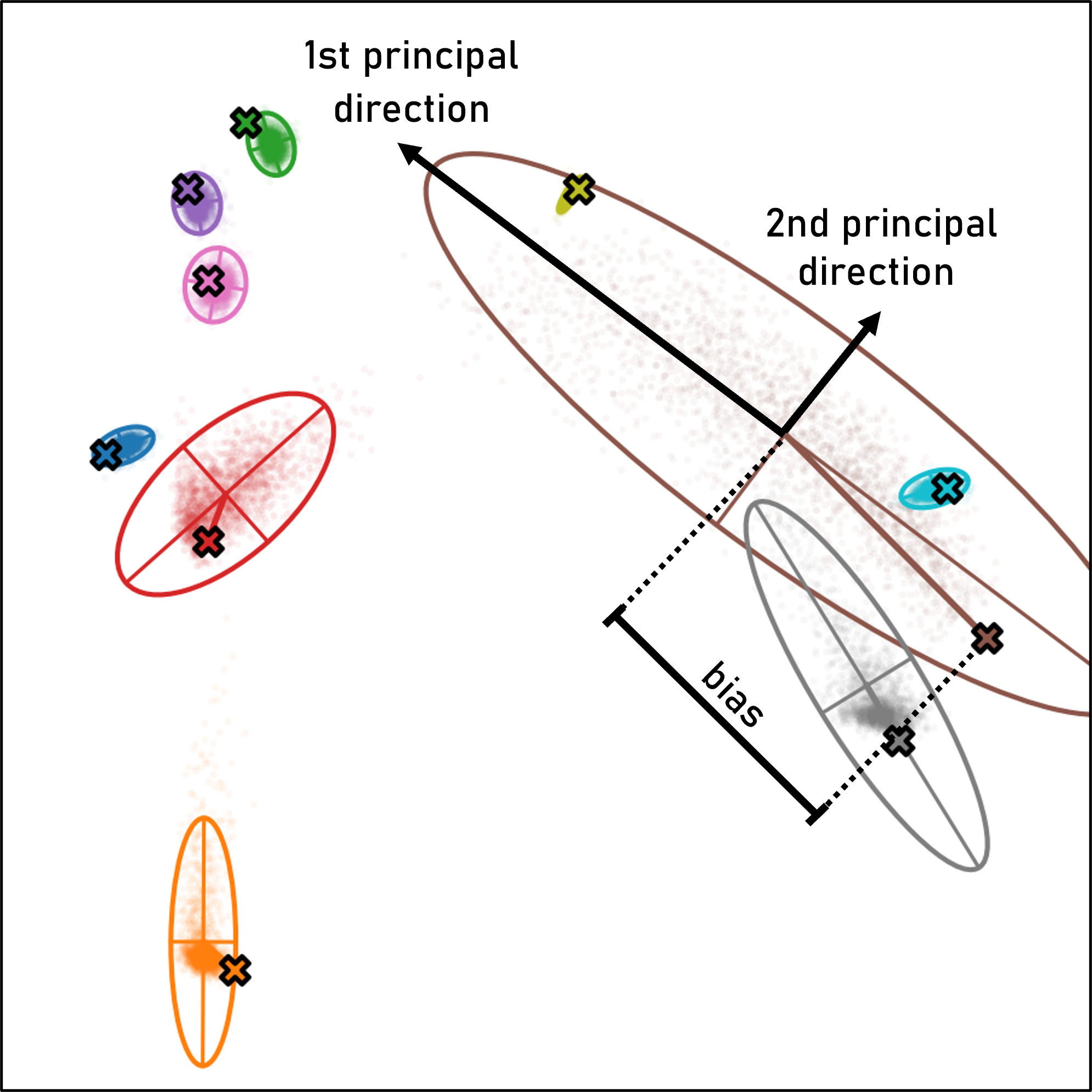}}
\end{wrapfigure}
\headit{Local PCA:} Local principal component analysis (PCA) captures the \emph{differential} structure of the projection around the anchor by approximating the local geometry of the noise-induced point cloud. The resulting principal directions and variances indicate how perturbations are amplified or attenuated along different directions in the projected space. This approach is inspired by Liao et al.~\cite{Liao2018ClusterBasedScatterplots}.
Changes in orientation or anisotropy provide evidence of deviations from locally linear noise propagation.
The local PCA ellipsoid is centered at the mean of the noise-induced projections and thus visually encodes the anchor's \emph{displacement bias} ($D_{\mathrm{bias}}$).
As an alternative, we propose Density Contours in the supplementary material.

\vspace{-0.4em}
\begin{wrapfigure}[12]{l}{3.8cm} % lines, orientation, size
    \vspace*{-3.5mm} % translation for margins
    %\hspace*{-4.5mm}
    \setlength{\fboxsep}{0pt}
    \setlength{\fboxrule}{0.6pt}
    \fbox{\includegraphics[width=4.3cm]{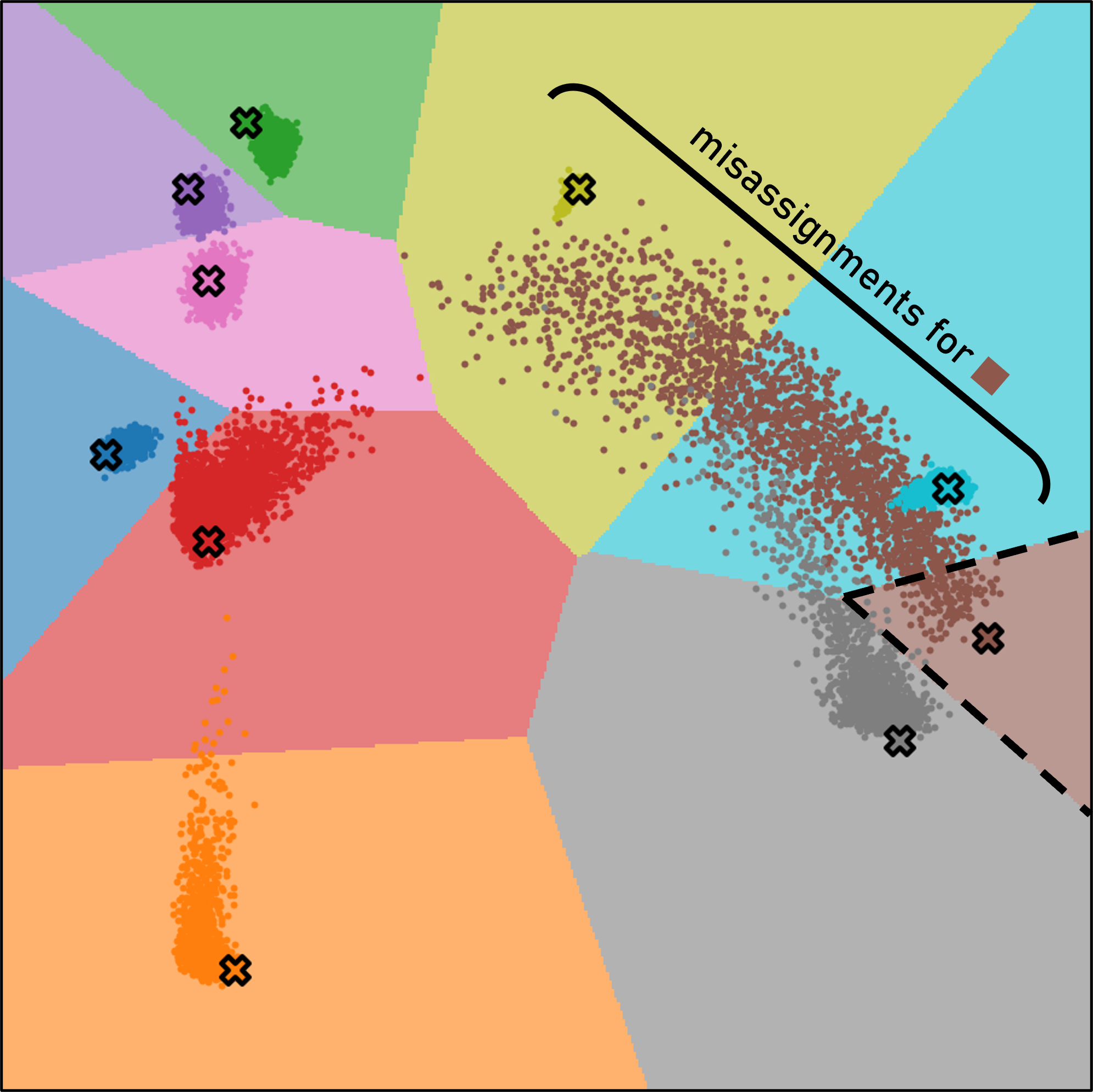}}
\end{wrapfigure}
\headit{Voronoi Assignment:}
This visualization introduces a \emph{comparative} perspective when multiple anchors are present.
By partitioning the projection space according to the nearest anchor, Voronoi regions~\cite{Aurenhammer1991} indicate which reference point dominates locally.
Points from one anchor's perturbation cloud that fall into a neighboring anchor's Voronoi cell signal potential misassignment under noise, identifying areas of reduced robustness with respect to competing anchors, visually linked to the \emph{nearest-anchor assignment error} ($E_{\mathrm{NA}}$).

\vspace{-1.2em}
\section{Evaluation}\label{sec:evaluation}

\headbf{Experimental Setup:} We use the two datasets in \autoref{tab:effective-r} and evaluate both UMAP and t-SNE as base projection methods~\cite{Blumberg2024}. We split each dataset using an 80-10-10 train-validation-test protocol.
Quality and stability metrics are computed on the held-out test set and averaged across 10 runs with different seeds (see \autoref{tab:comparison}).
Each run refits the base projection using the specified seed and uses independent network initializations as well as noise draws; anchors are class-centroid-based with count equal to the number of classes.
We report \emph{Trustworthiness} and \emph{Continuity} measures~\cite{Venna2001} as an average of all $T(k)$ with neighborhood size $k \in \{2, 4, 8, \dots, n / 2\}$ and $C(k)$ respectively~\cite{Colange2020}.
%For each point $i$, let $\rho_{ij}$ and $r_{ij}$ denote the rank of point $j$ in the high-dimensional and embedding spaces, respectively.
%$F_i(k)$ denotes the set of false neighbors of $i$ (points among the $k$ nearest in the embedding but not in the data space) and $M_i(k)$ the missed neighbors (points among the $k$ nearest in the data space but not in the embedding).
%Trustworthiness and Continuity are defined as
%$T(k)=1-\frac{1}{T_{\max}(k)}\sum_{i}\sum_{j\in F_i(k)}(\rho_{ij}-k),
%$
%and
%$
%C(k)=1-\frac{1}{C_{\max}(k)}\sum_{i}\sum_{j\in M_i(k)}(r_{ij}-k),
%$
%where the normalization constants $T_{\max}(k)=C_{\max}(k)$ ensure values lie in $[0,1]$ and are equal to
%$k N(2N-3k-1)/2$ for $k<N/2$, or $N(N-k)(N-k-1)/2$ for $k>N/2$.
Higher $T$ and $C$ indicate better local structure preservation; results are in \autoref{tab:comparison}.

%\vspace{-0.25em}
\headbf{Neural Network Architectures and Training:}
We train fully connected MLPs with ReLU activations learning $f: \mathbb{R}^d \to \mathbb{R}^2$.
\texttt{MLP-small}: 3 hidden layers, 512 units each;
\texttt{MLP-large}: 6 hidden layers, 1024 units each.
Our general objective function is $\mathcal{L}_{\mathrm{MSE}} = \mathbb{E}_{(x,y)}\!\left[ \left\lVert f(x) - y \right\rVert_2^2 \right]$~\cite{Espadoto2020Deep,Dennig2025Evaluating}.
We use the Adam optimizer~\cite{Kingma2015}, batch size 256, learning rate $10^{-3}$, and 100 epochs with early stopping (patience=10).
No noise augmentation is applied during training.
The \texttt{+J} suffix denotes the application of Jacobian regularization. It penalizes the Frobenius norm of the input-output Jacobian $J_f(x) = \partial f_\theta(x)/\partial x \in \mathbb{R}^{k \times d}$, adding
$\mathcal{L}_{\mathrm{Jac}} = \lambda \, \mathbb{E}_{x}\bigl[ \lVert J_f(x) \rVert_F^2 \bigr]$
to the training loss (we choose $\lambda{=}10$; see sweep over $\lambda$ in supplementary material), encouraging locally smooth mappings that are more robust to small input perturbations~\cite{Jakubovitz2018}.

\begin{table}[t]
\setlength{\tabcolsep}{1.1mm}
\small
\centering
\begin{tabular}{lcccc}
\textbf{Model} & \shortstack{\textbf{MNIST}\\\textbf{(UMAP)}} & \shortstack{\textbf{Fashion}\\\textbf{(UMAP)}} & \shortstack{\textbf{MNIST}\\\textbf{(t-SNE)}} & \shortstack{\textbf{Fashion}\\\textbf{(t-SNE)}} \\
\hline
\hline
\multicolumn{5}{c}{\textit{Average MSE Loss (lower is better)}} \\
\hline
\texttt{MLP-small} & .829 $\pm$ .104 & .293 $\pm$ .054 & \underline{62.7 $\pm$ 3.8\phantom{0}} & 30.7 $\pm$ 2.98 \\
\texttt{MLP-small+J} & .676 $\pm$ .035 & \textbf{.279 $\pm$ .046} & 61.8 $\pm$ 2.15 & 29.9 $\pm$ 2.81 \\
\texttt{MLP-large} & \underline{.855 $\pm$ .108} & \underline{.33\phantom{0} $\pm$ .044} & 52.7 $\pm$ 6.34 & \underline{32\phantom{0} $\pm$ 3.34} \\
\texttt{MLP-large+J} & \textbf{.668 $\pm$ .06\phantom{0}} & .297 $\pm$ .049 & \textbf{48.8 $\pm$ 4.9\phantom{0}} & \textbf{26.1 $\pm$ 1.71} \\
\hline
\multicolumn{5}{c}{\textit{Avg. Trustworthiness $T(k)$ with $k \in \{2, 4, 8, \dots, n / 2\}$ (higher is better)}} \\
\hline
\texttt{MLP-small} & .859 $\pm$ .004 & .928 $\pm$ .002 & .857 $\pm$ .001 & .928 $\pm$ .001 \\
\texttt{MLP-small+J} & \underline{.858 $\pm$ .002} & \underline{.927 $\pm$ .002} & \underline{.85\phantom{0} $\pm$ .001} & \underline{.926 $\pm$ .001} \\
\texttt{MLP-large} & \textbf{.867 $\pm$ .003} & \textbf{.929 $\pm$ .002} & \textbf{.872 $\pm$ .001} & \textbf{.931 $\pm$ .001} \\
\texttt{MLP-large+J} & .865 $\pm$ .001 & .929 $\pm$ .002 & .863 $\pm$ .001 & .929 $\pm$ .001 \\
\hline
\multicolumn{5}{c}{\textit{Avg. Continuity $C(k)$ with $k \in \{2, 4, 8, \dots, n / 2\}$ (higher is better)}} \\
\hline
\texttt{MLP-small} & .864 $\pm$ .001 & .938 $\pm$ .002 & \underline{.871 $\pm$ .001} & .943 $\pm$ .001 \\
\texttt{MLP-small+J} & \textbf{.866 $\pm$ .001} & \textbf{.939 $\pm$ .002} & \textbf{.873 $\pm$ .001} & \textbf{.944 $\pm$ .001} \\
\texttt{MLP-large} & \underline{.864 $\pm$ .002} & \underline{.938 $\pm$ .002} & .871 $\pm$ .001 & \underline{.942 $\pm$ .001} \\
\texttt{MLP-large+J} & .866 $\pm$ .001 & .939 $\pm$ .002 & .872 $\pm$ .001 & .943 $\pm$ .001 \\
\hline
\multicolumn{5}{c}{\textit{Mean Displacement (lower is better)}} \\
\hline
\texttt{MLP-small} & \underline{1.8\phantom{0} $\pm$ .382} & \underline{.863 $\pm$ .175} & \underline{24.8 $\pm$ 1.59} & \underline{11.7 $\pm$ 2.69} \\
\texttt{MLP-small+J} & .265 $\pm$ .051 & .465 $\pm$ .204 & 5.03 $\pm$ .612 & 6.06 $\pm$ 1.38 \\
\texttt{MLP-large} & .562 $\pm$ .282 & .665 $\pm$ .244 & 6.34 $\pm$ 1.41 & 10.7 $\pm$ 2.55 \\
\texttt{MLP-large+J} & \textbf{.207 $\pm$ .069} & \textbf{.391 $\pm$ .091} & \textbf{4.58 $\pm$ 1.4\phantom{0}} & \textbf{5.35 $\pm$ 1.36} \\
\hline
\multicolumn{5}{c}{\textit{Displacement Bias (lower is better)}} \\
\hline
\texttt{MLP-small} & \underline{1.76 $\pm$ .375} & \underline{.772 $\pm$ .179} & \underline{24.4 $\pm$ 1.62} & \underline{10.4 $\pm$ 2.74} \\
\texttt{MLP-small+J} & .233 $\pm$ .053 & .435 $\pm$ .208 & 4.53 $\pm$ .617 & 5.47 $\pm$ 1.37 \\
\texttt{MLP-large} & .524 $\pm$ .282 & .622 $\pm$ .246 & 5.35 $\pm$ 1.63 & 9.67 $\pm$ 2.65 \\
\texttt{MLP-large+J} & \textbf{.18\phantom{0} $\pm$ .066} & \textbf{.361 $\pm$ .093} & \textbf{4.13 $\pm$ 1.41} & \textbf{4.78 $\pm$ 1.34} \\
\hline
\multicolumn{5}{c}{\textit{Average Nearest-Anchor Assignment Error (lower is better)}} \\
\hline
\texttt{MLP-small} & \underline{.294 $\pm$ .069} & \underline{.109 $\pm$ .038} & \underline{.307 $\pm$ .075} & \underline{.192 $\pm$ .081} \\
\texttt{MLP-small+J} & .006 $\pm$ .012 & .037 $\pm$ .039 & \textbf{.017 $\pm$ .022} & .068 $\pm$ .053 \\
\texttt{MLP-large} & .069 $\pm$ .06\phantom{0} & .078 $\pm$ .047 & .024 $\pm$ .022 & .164 $\pm$ .091 \\
\texttt{MLP-large+J} & \textbf{.003 $\pm$ .009} & \textbf{.031 $\pm$ .034} & .019 $\pm$ .031 & \textbf{.036 $\pm$ .046} \\
\end{tabular}
\caption{Stability and quality metrics for MLP-based parametric projections based on UMAP and t-SNE with different regularization strategies across datasets. Bold values indicate best, underlined values worst performance per metric and dataset.}
\label{tab:comparison}
\vspace{-2em}
\end{table}

%\vspace{-0.25em}
\headbf{Quantitative Results:}
\autoref{tab:comparison} summarizes six metrics across two datasets and two projection methods (UMAP, t-SNE), averaged over 10 runs.
Jacobian regularization consistently improves stability for both architectures and both projection methods: \texttt{MLP-large+J} achieves the lowest mean displacement in all settings, reducing displacement by 28--63\% relative to its unregularized counterpart \texttt{MLP-large}.
The effect is equally pronounced for the smaller network: on \emph{MNIST} (UMAP), \texttt{MLP-small+J} reduces mean displacement from 1.80 to .265 ($-85\%$) and anchor assignment error from .294 to .006.
The same pattern holds for t-SNE: on \emph{MNIST} (t-SNE), \texttt{MLP-small+J} reduces displacement from 24.8 to 5.03 ($-80\%$) and anchor error from .307 to .017.
Notably, on \emph{MNIST} (UMAP), the unregularized \texttt{MLP-small} exhibits mean displacement (1.80) nearly equal to displacement bias (1.76), indicating systematic drift rather than isotropic spread; the same effect appears on \emph{MNIST} (t-SNE) with displacement 24.8 vs.\ bias 24.4.
Increasing network capacity alone provides moderate stability gains (\texttt{MLP-large} vs.\ \texttt{MLP-small}), but Jacobian regularization is substantially more effective: On \emph{Fashion} (UMAP), \texttt{MLP-large}'s displacement drops to .665, whereas adding \texttt{+J} further lowers it to .391.
Reconstruction loss and stability do not trade off uniformly; \texttt{MLP-large+J} achieves the best MSE on \emph{MNIST} (UMAP) (.668) while simultaneously being the most stable.
The averaged \emph{Trustworthiness} and \emph{Continuity} remain near-constant across all methods ($T$: .850--.931, $C$: .864--.944), confirming that neighborhood-preservation metrics fail to capture the stability differences revealed by displacement-based metrics, i.e., the failure mode anticipated in \autoref{sec:introduction}.
Our stability metrics are cheap to compute.
The total wall-clock per model on \emph{MNIST}/\emph{Fashion} is ${\sim}135$\,ms for \texttt{MLP-large}, dominated by MLP forward passes; $E_\mathrm{NA}$ is $O(NA^2)$ and sub-millisecond.

\begin{figure}
\newlength{\imgwidth}
\setlength{\imgwidth}{0.28\linewidth}
\setlength{\tabcolsep}{1.0mm}
\setlength{\fboxsep}{0pt}
\setlength{\fboxrule}{0.6pt}
\newcommand{\compimg}[1]{%
  {\includegraphics[width=\imgwidth]{#1}}%
}
\newcommand{\ph}{\vphantom{Mg}}
\small
\centering
\begin{tabular}{lccc}
 & \textbf{Reference Projection} & \horizontalLabel{\imgwidth}{Most Stable\\[-0.2em]{\normalfont\texttt{MLP-large+J}}\\[0.3em]} & \horizontalLabel{\imgwidth}{Least Stable\\[-0.2em]{\normalfont\texttt{MLP-small}}\\[0.3em]} \\
%
% --- MNIST (UMAP): Anchor Lines ---
\verticalLabel{\imgwidth}{MNIST (UMAP)\\[-0.2em]{\normalfont \emph{Anchor Lines}}\\[-0.3em]} &
  \fbox{\compimg{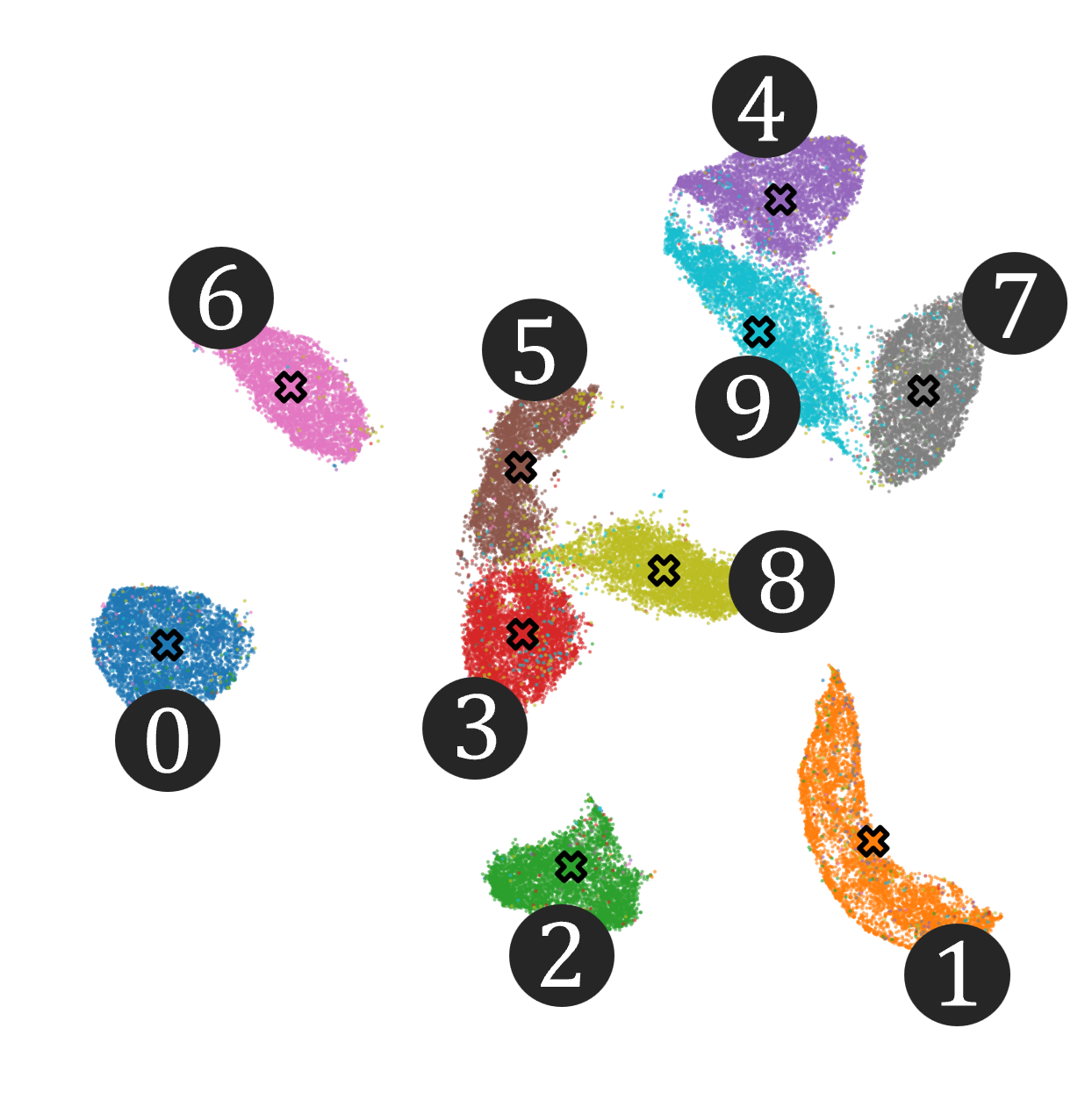}} &
  \fbox{\compimg{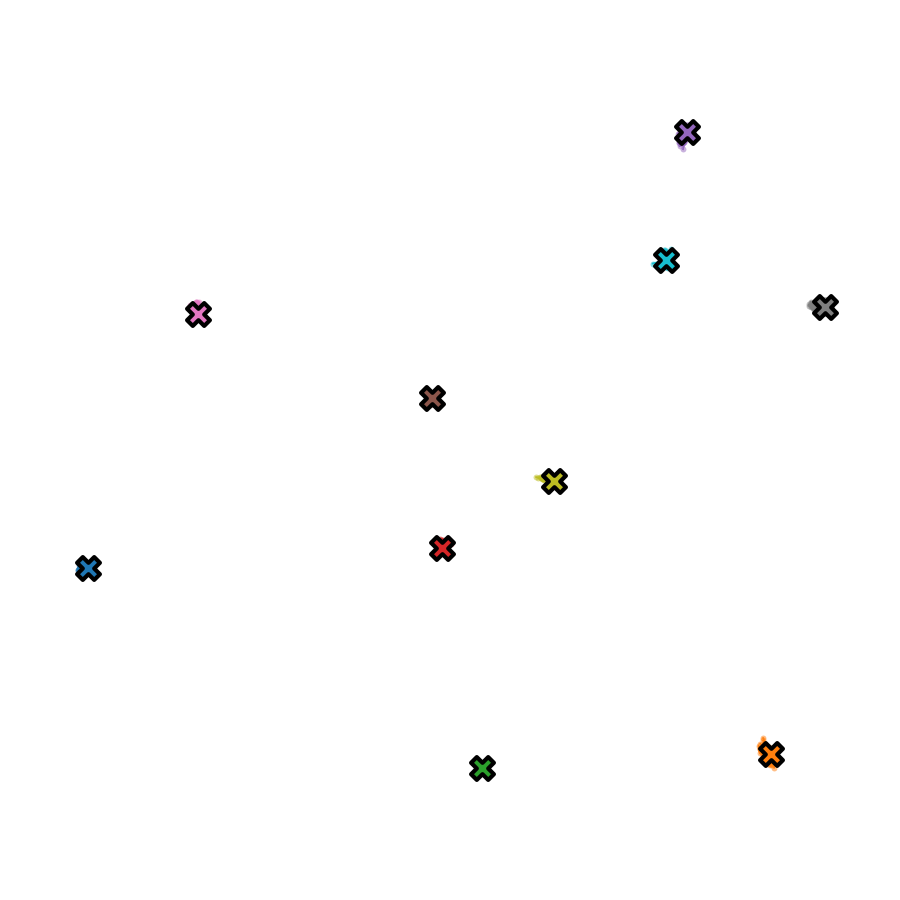}} &
  \fbox{\compimg{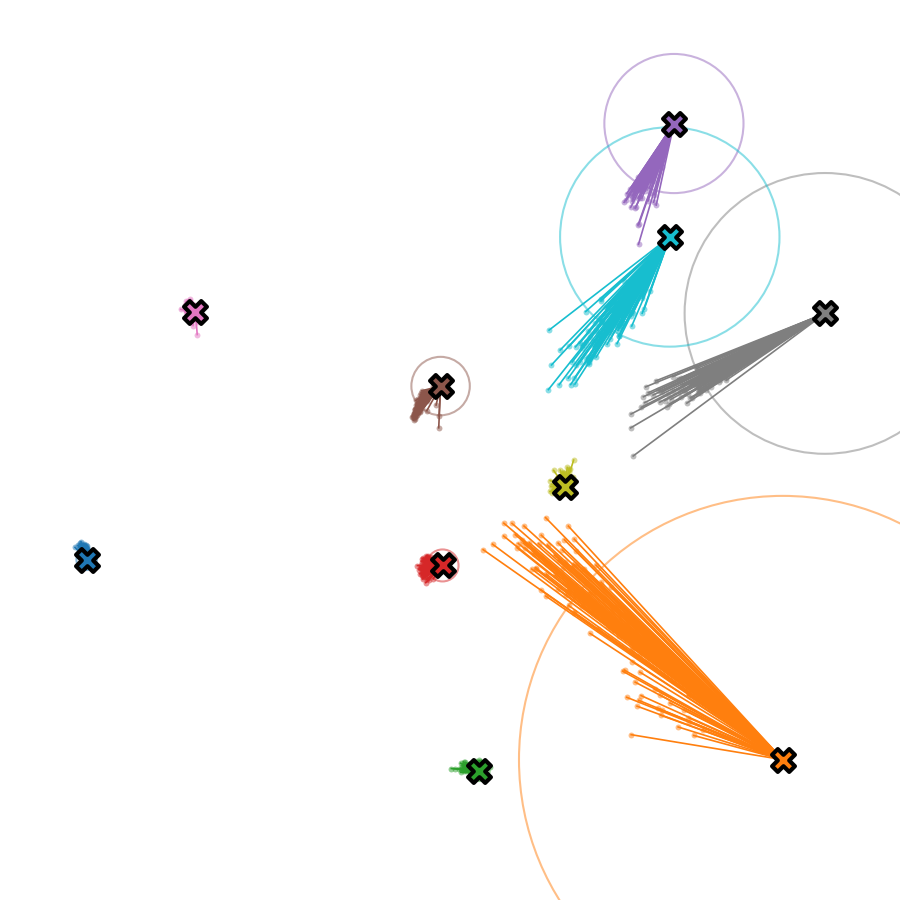}} \\[0.5em]
%
% --- Fashion (UMAP): Local PCA ---
\verticalLabel{\imgwidth}{Fashion (UMAP)\\[-0.2em]{\normalfont \emph{Local PCA}}\\[-0.3em]} &
  \fbox{\compimg{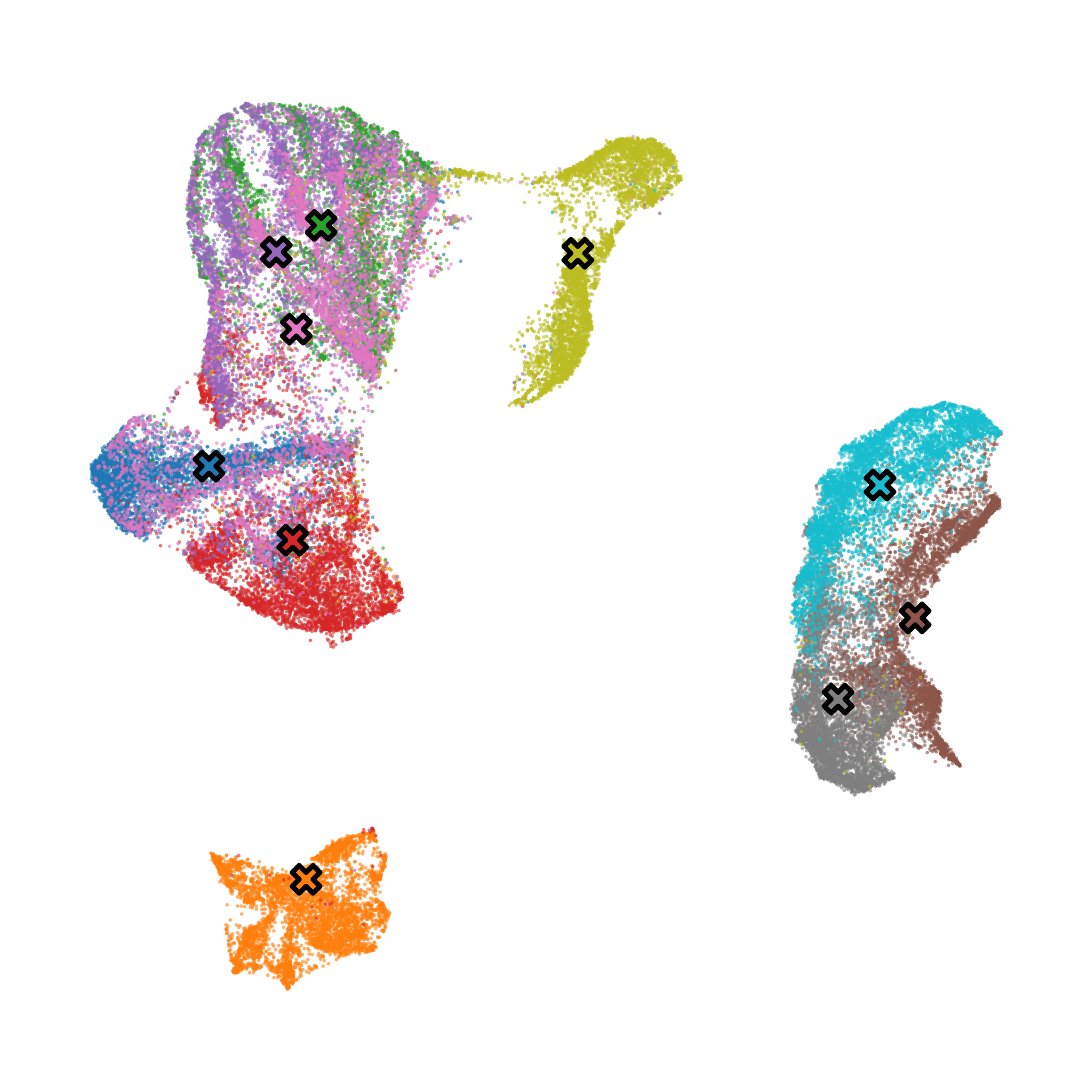}} &
  \fbox{\compimg{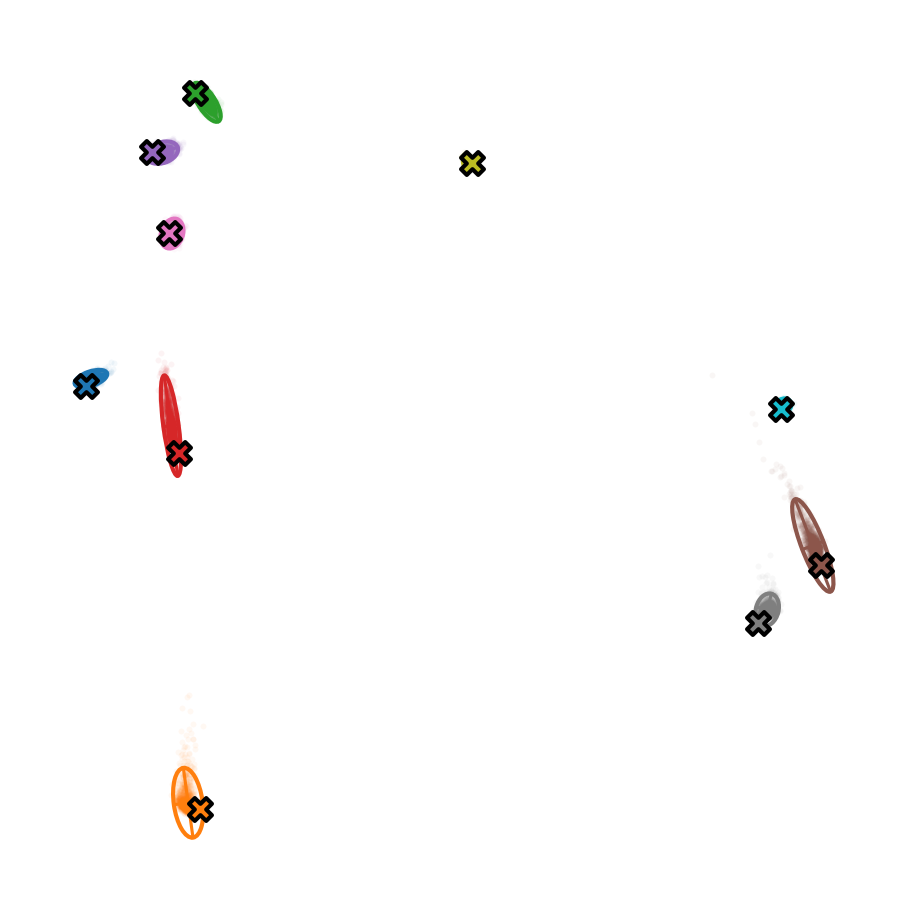}} &
  \fbox{\compimg{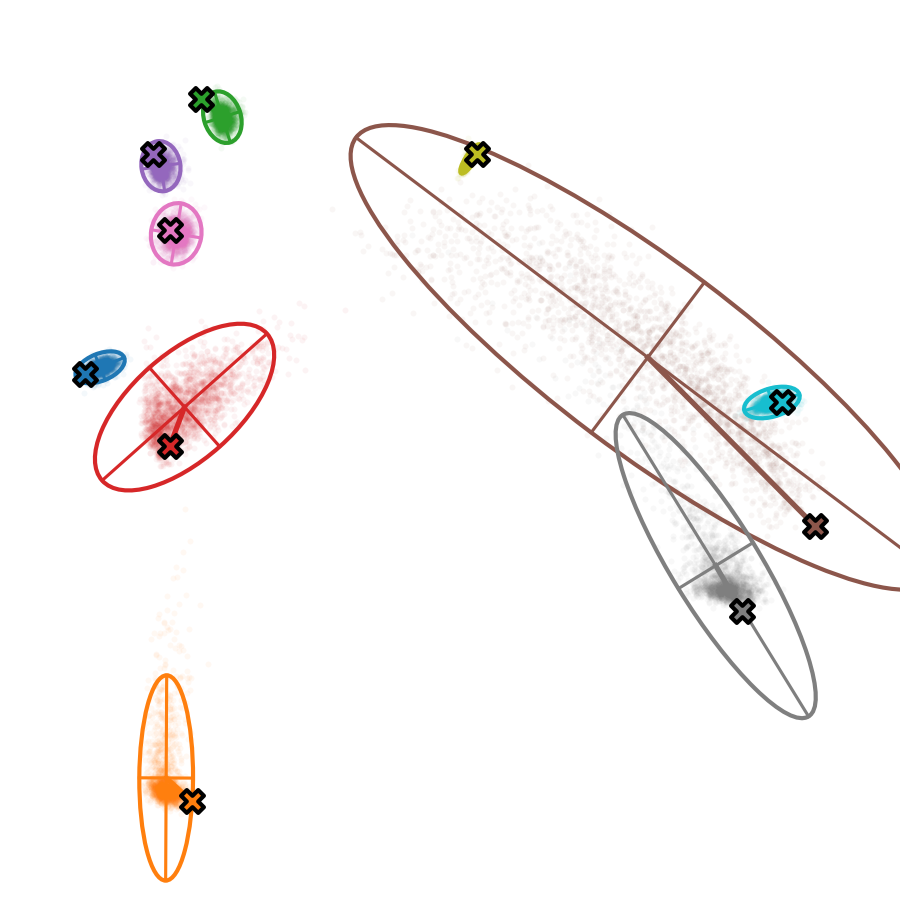}} \\[0.5em]
%
% --- MNIST (t-SNE): Voronoi ---
\verticalLabel{\imgwidth}{MNIST (t-SNE)\\[-0.2em]{\normalfont \emph{Voronoi Assignment}}\\[-0.3em]} &
  \fbox{\compimg{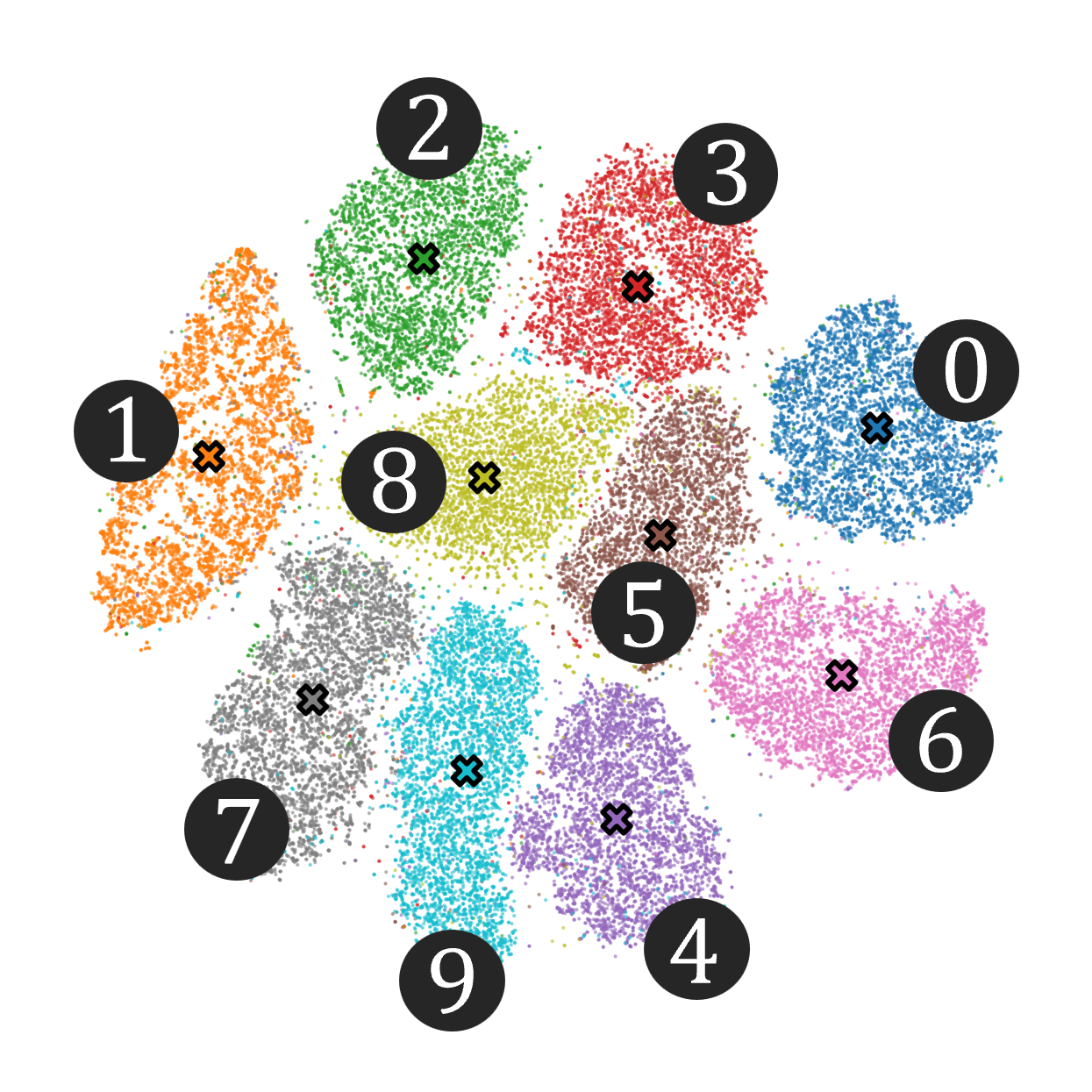}} &
  \fbox{\compimg{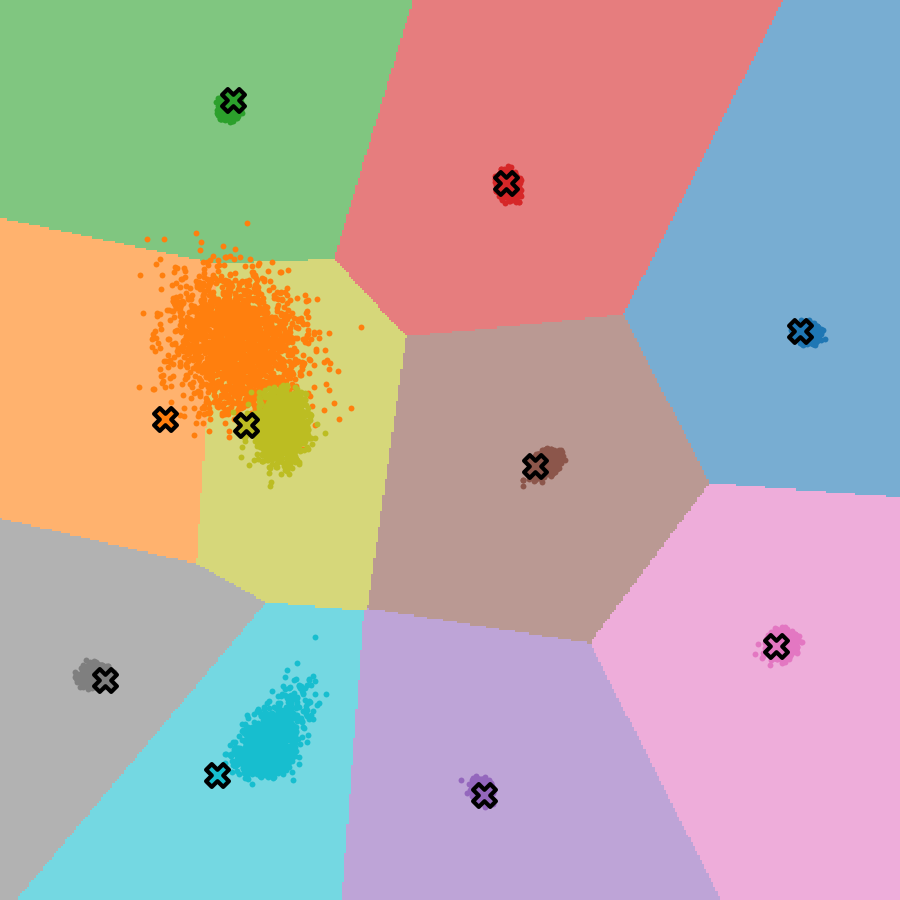}} &
  \fbox{\compimg{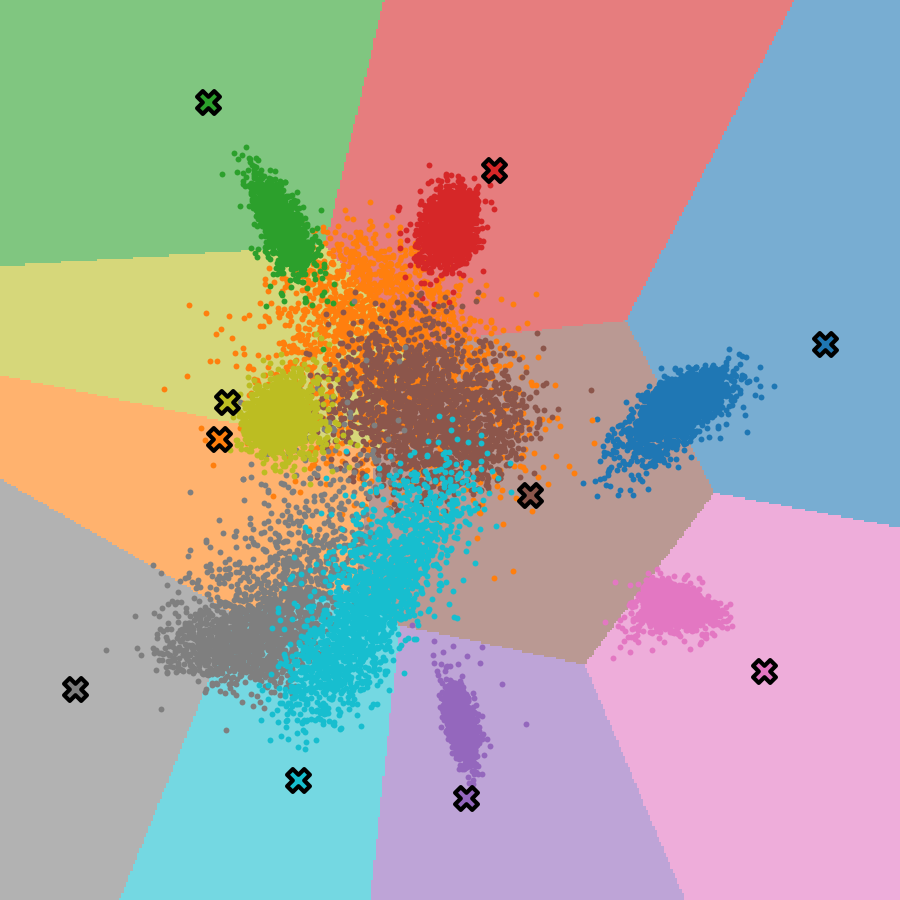}} \\[0.5em]
%
% --- Fashion (t-SNE): Anchor Lines ---
\verticalLabel{\imgwidth}{Fashion (t-SNE)\\[-0.2em]{\normalfont \emph{Anchor Lines}}\\[-0.3em]} &
  \fbox{\compimg{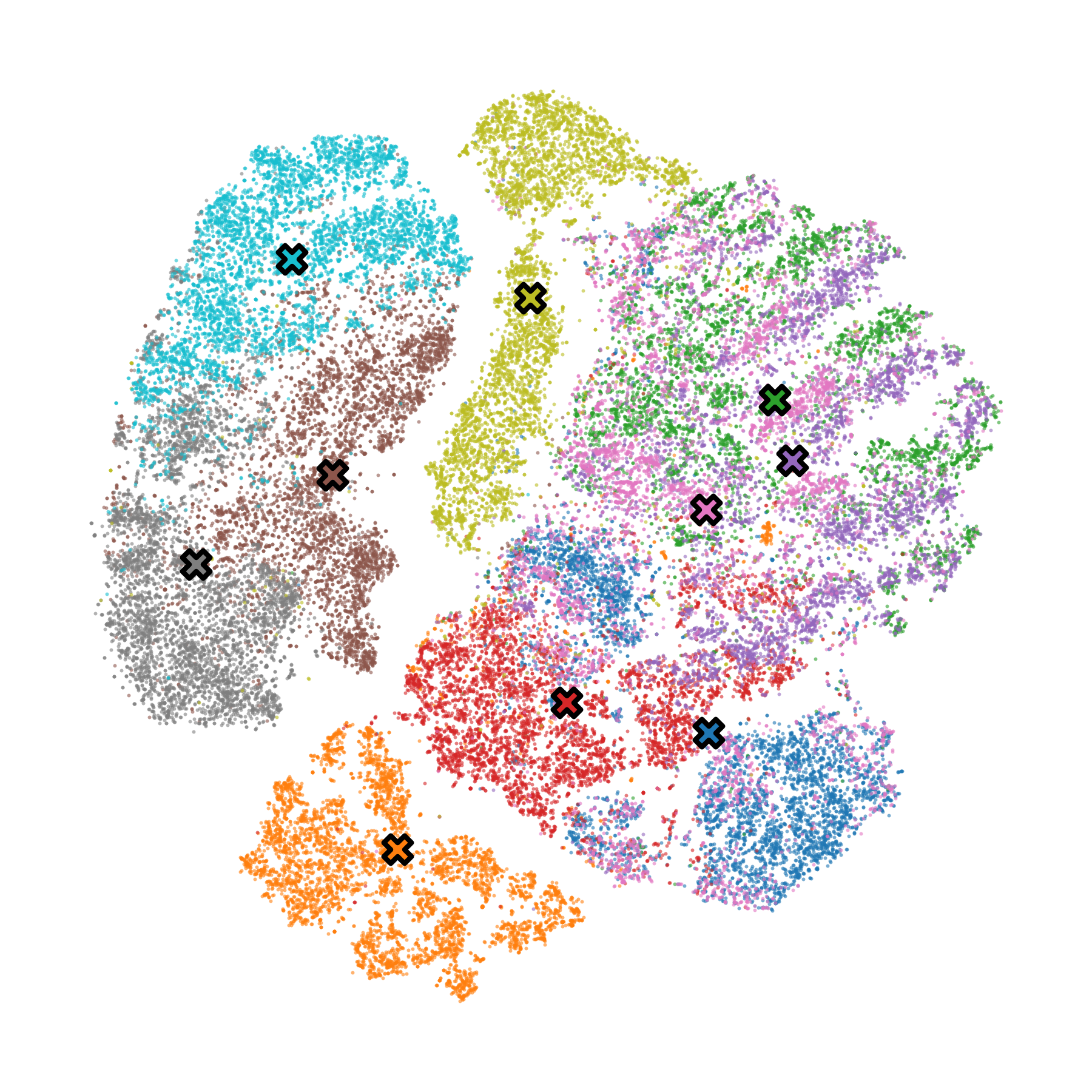}} &
  \fbox{\compimg{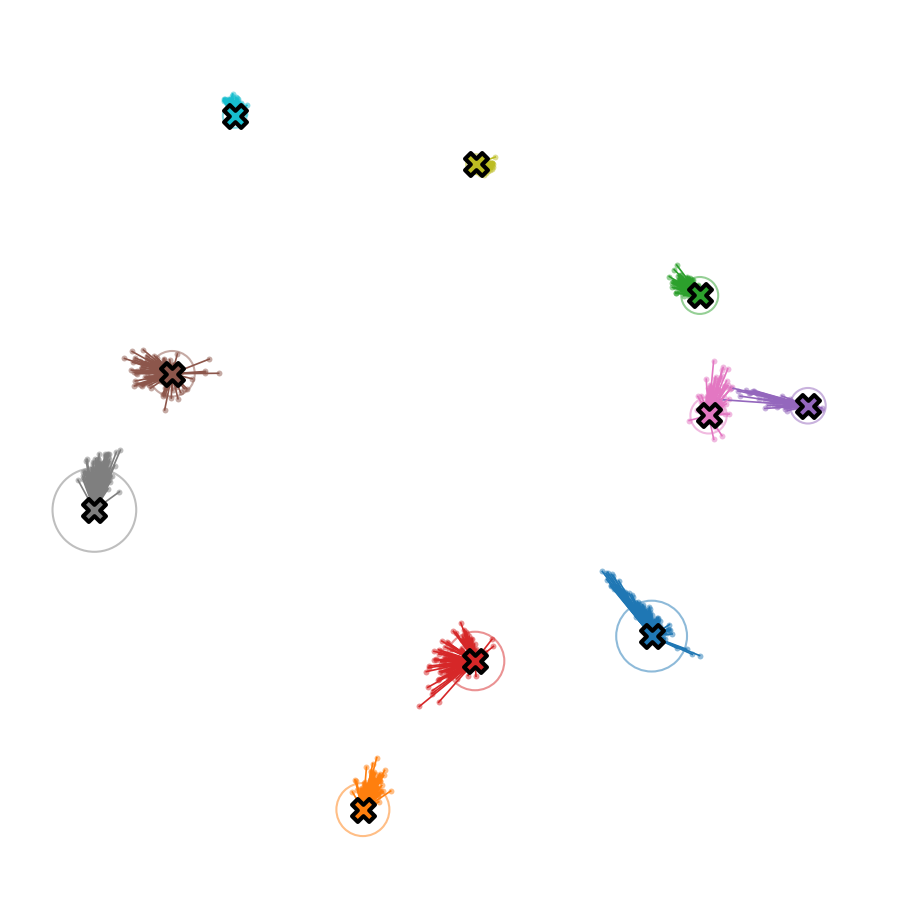}} &
  \fbox{\compimg{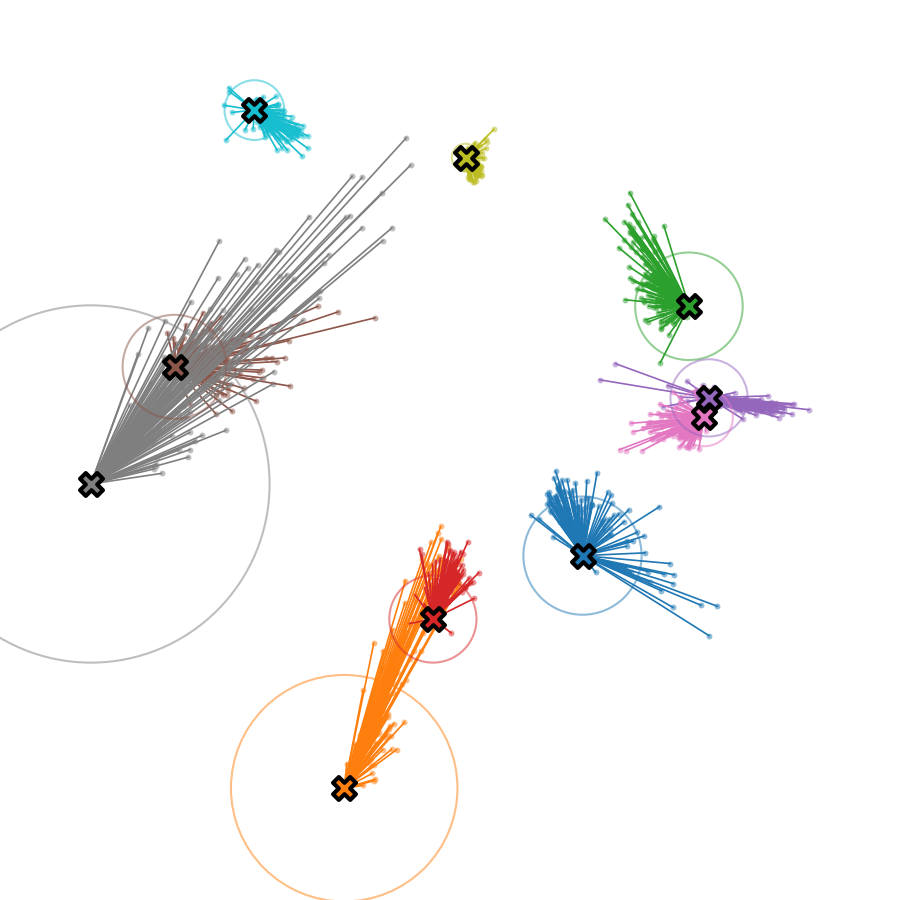}} \\
\end{tabular}
\caption{Qualitative stability comparison across UMAP and t-SNE projections. Each row shows the reference projection with anchors, the most and least stable MLP by mean displacement (\autoref{tab:comparison}) with stability visualizations.}
\label{fig:comparison}
\vspace{-2em}
\end{figure}

\vspace{-0.25em}
\headbf{Qualitative Results:}
\autoref{fig:comparison} contrasts the most (\texttt{MLP-large+J}) and least (\texttt{MLP-small}) stable models per setting by mean displacement, each row using a different stability visualization alongside the reference projection.
On \emph{MNIST} (UMAP), anchor lines for \texttt{MLP-large+J} remain tightly clustered around anchors, while \texttt{MLP-small} shows wide displacement fans.
For \emph{Fashion} (UMAP), local PCA ellipses reveal both larger extent and stronger directional anisotropy for \texttt{MLP-small}, consistent with its higher displacement bias.
The \emph{MNIST} (t-SNE) Voronoi row provides the starkest contrast: \texttt{MLP-large+J} confines perturbed points within their anchor's Voronoi cell (nearest-anchor assignment error .078), while \texttt{MLP-small} scatters points across cell boundaries (.329), with displacement nearly entirely due to systematic bias.
On \emph{Fashion} (t-SNE), shorter anchor lines confirm reduced displacement for \texttt{MLP-large+J} relative to \texttt{MLP-small}.
%Across all rows, the visualizations reveal stability differences that the near-constant \emph{Trustworthiness} and \emph{Continuity} scores in \autoref{tab:comparison} cannot capture on their own.

\vspace{-1em}
\section{Discussion}

Our evaluation yields two main findings:
(1)~Standard neighborhood-based quality metrics (Trustworthiness, Continuity) remain virtually identical across all tested methods ($T$: .850--.931, $C$: .864--.944), failing to reflect substantial stability differences. Our displacement-based measures and visual diagnostics close this gap, revealing instabilities that $T$ and $C$ cannot capture.
(2)~As a practical demonstration, Jacobian regularization, with an expected but now quantifiable effect, reduces displacement by up to 85\% (UMAP) and 80\% (t-SNE), and anchor assignment error to near zero on \emph{MNIST}, with consistent gains across both projection methods and datasets. Comparing two network sizes with and without regularization confirms that Jacobian regularization is more effective than increasing capacity alone.
We also test spectral normalization and report results on two additional datasets in the supplementary material.

\vspace{-0.25em}
\headbf{Implications for Visual Analytics:}
The stability of MLP projections directly determines whether analysts can trust that spatial patterns represent genuine data structures~\cite{Nonato2019}.
Our evaluation shows that standard neighborhood-preservation metrics are insufficient: Methods with near-identical Trustworthiness and Continuity scores exhibit displacement differences exceeding 80\%.
Our visual diagnostics let practitioners inspect \emph{where} and \emph{in which direction} a projection is sensitive.
Our findings indicate that regularization strategy matters more than network capacity: A compact Jacobian-regularized network achieves stability at lower computational cost (training times are reported in the supplementary material).
Overall, we advocate for incorporating stability assessment when training MLP-based parametric projections.

\vspace{-0.25em}
\headbf{Limitations and Future Work:}
Jacobian regularization minimizes local sensitivity; displacement measures quantify local sensitivity. Hence, the stability improvement is expected.
We analyze the cost-benefit tradeoff (stability gain vs. MSE/T/C cost) rather than an unexpected effect.
Our measurements characterize the learned parametric mapping, not the base DR, and do not disentangle NN- from DR-induced effects.
Isotropic Gaussian noise may not match realistic perturbation distributions for image data, though in our study it serves as a proxy for sensor drift; more realistic perturbations could be obtained by interpolating between same-class samples or via inverse projection~\cite{Espadoto2021Unprojection,Dennig2025Evaluating}.
Our measures are defined as functions of $\sigma$, but we evaluate only a single noise level per dataset; a sweep across $\sigma$ values would provide a more complete stability profile. We select $\sigma$ in a dataset-adaptive manner via the 0.25th percentile of pairwise distances (\autoref{tab:effective-r}), ensuring perturbations remain within a realistic range for each dataset.
We evaluate only MLP architectures; extending the framework to AEs, CNNs, or other parametric DR variants would broaden its applicability.
Only class-centroid-based anchors are tested; thus, an analysis of other anchor selection strategies would strengthen the evaluation.

\vspace{-1em}
\section{Conclusion}

Parametric projections promise real-time inference and out-of-sample extension, yet standard quality metrics ignore their sensitivity to input perturbations.
We introduced displacement-based stability measures and visualizations that reveal instabilities invisible to \emph{Trustworthiness} and \emph{Continuity} measures.
Jacobian regularization can mitigate such instability, as confirmed by our diagnostic measures, at minimal cost to reconstruction fidelity.
The takeaway is concrete: Pair parametric projections with smoothness constraints and validate stability before deployment in noise-prone scenarios.

\headbf{Acknowledgments:}
This work was funded by the Deutsche Forschungsgemeinschaft (DFG, German Research Foundation) -- Project-ID 251654672 -- TRR 161 (Project A03).

%-------------------------------------------------------------------------
% bibtex
\bibliographystyle{eg-alpha-doi}  
\bibliography{references}        

@article{Hinterreiter2023,
  author       = {Andreas P. Hinterreiter and
                  Christina Humer and
                  Bernhard Kainz and
                  Marc Streit},
  title        = {{ParaDime}: {A} Framework for Parametric Dimensionality Reduction},
  journal      = {Comput. Graph. Forum},
  volume       = {42},
  number       = {3},
  pages        = {337--348},
  year         = {2023},
  _url         = {https://doi.org/10.1111/cgf.14834},
  doi          = {10.1111/CGF.14834}
}

@article{Sainburg2021,
  author       = {Tim Sainburg and
                  Leland McInnes and
                  Timothy Q. Gentner},
  title        = {Parametric {UMAP} Embeddings for Representation and Semisupervised
                  Learning},
  journal      = {Neural Comput.},
  volume       = {33},
  number       = {11},
  pages        = {2881--2907},
  year         = {2021},
  _url         = {https://doi.org/10.1162/neco\_a\_01434},
  doi          = {10.1162/NECO\_A\_01434}
}

@misc{Lecun1998,
  title     = {{MNIST} handwritten digit database},
  author    = {LeCun, Yann and Cortes, Corinna and Burges, Chris},
  year      = {1998},
  _url      = {http://yann.lecun.com/exdb/mnist/}
}

@article{Xiao2017,
  author       = {Han Xiao and
                  Kashif Rasul and
                  Roland Vollgraf},
  title        = {{Fashion-MNIST}: {A} Novel Image Dataset for Benchmarking Machine Learning
                  Algorithms},
  journal      = {CoRR},
  volume       = {abs/1708.07747},
  year         = {2017},
  _url         = {http://arxiv.org/abs/1708.07747},
  eprinttype   = {arXiv},
  eprint       = {1708.07747}
}

@inproceedings{Dennig2025Evaluating,
  author     = {Dennig, Frederik L. and Geyer, Nina and Blumberg, Daniela and Metz, Yannick and Keim, Daniel A.},
  _editor    = {Schulz, Hans-Jörg and Villanova, Anna},
  title      = {{Evaluating Autoencoders for Parametric and Invertible Multidimensional Projections}},
  booktitle  = {16th Int. EuroVis Workshop Vis. Anal.},
  _publisher = {The Eurographics Association},
  year       = {2025},
  doi        = {10.2312/eurova.20251099},
  _url       = {https://diglib.eg.org/items/123db003-3312-4e4c-ba14-0c910}
}

@inproceedings{Blumberg2024,
  booktitle  = {15th Int. EuroVis Workshop Vis. Anal.},
  _editor    = {El-Assady, Mennatallah and Schulz, Hans-Jörg},
  title      = {{Inverting Multidimensional Scaling Projections Using Data Point Multilateration}},
  author     = {Blumberg, Daniela and Wang, Yu and Telea, Alexandru and Keim, Daniel A. and Dennig, Frederik L.},
  year       = {2024},
  _publisher = {The Eurographics Association},
  _isbn      = {978-3-03868-253-0},
  doi        = {10.2312/eurova.20241112}
}

@article{Espadoto2020Deep,
  author       = {Mateus Espadoto and
                  Nina Sumiko Tomita Hirata and
                  Alexandru C. Telea},
  title        = {Deep learning multidimensional projections},
  journal      = {Inf. Vis.},
  volume       = {19},
  number       = {3},
  pages        = {247--269},
  year         = {2020},
  _url         = {https://doi.org/10.1177/1473871620909485},
  doi          = {10.1177/1473871620909485}
}

@inproceedings{Maaten2009,
  author       = {Laurens van der Maaten},
  _editor      = {David A. Van Dyk and
                  Max Welling},
  title        = {Learning a Parametric Embedding by Preserving Local Structure},
  booktitle    = {12th Int. Conf. Artif. Intell. Stat.},
  _series      = {{JMLR} Proceedings},
  volume       = {5},
  pages        = {384--391},
  _publisher   = {JMLR.org},
  year         = {2009},
  _url         = {http://proceedings.mlr.press/v5/maaten09a.html},
}

@article{Maaten2008Tsne,
  author  = {Laurens van der Maaten and Geoffrey Hinton},
  title   = {Visualizing Data using {t-SNE}},
  journal = {J. Mach. Learn. Res.},
  year    = {2008},
  volume  = {9},
  number  = {86},
  pages   = {2579--2605},
  _url    = {http://jmlr.org/papers/v9/vandermaaten08a.html}
}

@article{Kruskal1978Multidimensional,
  title      = {Multidimensional Scaling},
  author     = {Kruskal, Joseph and Wish, Myron},
  journal    = {Murry Hill},
  year       = {1978},
  _publisher = {SAGE Publications, Inc.},
  doi        = {10.4135/9781412985130},
  _url       = {https://methods.sagepub.com/book/mono/multidimensional-scaling/toc}
}

@article{McInnes2018,
  author       = {Leland McInnes and
                  John Healy and
                  Nathaniel Saul and
                  Lukas Grossberger},
  title        = {{UMAP: Uniform Manifold Approximation and Projection}},
  journal      = {J. Open Source Softw.},
  volume       = {3},
  number       = {29},
  pages        = {861},
  year         = {2018},
  doi          = {10.21105/joss.00861}
}

@article{Bunte2012,
  author       = {Kerstin Bunte and
                  Michael Biehl and
                  Barbara Hammer},
  title        = {A General Framework for Dimensionality-Reducing Data Visualization
                  Mapping},
  journal      = {Neural Comput.},
  volume       = {24},
  number       = {3},
  pages        = {771--804},
  year         = {2012},
  _url         = {https://doi.org/10.1162/NECO\_a\_00250},
  doi          = {10.1162/NECO\_A\_00250}
}

@article{Appleby2022,
  author       = {Gabriel Appleby and
                  Mateus Espadoto and
                  Rui Chen and
                  Samuel Goree and
                  Alexandru C. Telea and
                  Erik W. Anderson and
                  Remco Chang},
  title        = {{HyperNP}: Interactive Visual Exploration of Multidimensional Projection
                  Hyperparameters},
  journal      = {Comput. Graph. Forum},
  volume       = {41},
  number       = {3},
  pages        = {169--181},
  year         = {2022},
  _url         = {https://doi.org/10.1111/cgf.14531},
  doi          = {10.1111/CGF.14531}
}

@article{Nonato2019,
  author       = {Luis Gustavo Nonato and
                  Micha{\"{e}}l Aupetit},
  title        = {Multidimensional Projection for Visual Analytics: Linking Techniques
                  with Distortions, Tasks, and Layout Enrichment},
  journal      = {{IEEE} Trans. Vis. Comput. Graph.},
  volume       = {25},
  number       = {8},
  _pages       = {2650--2673},
  year         = {2019},
  _url         = {https://doi.org/10.1109/TVCG.2018.2846735},
  doi          = {10.1109/TVCG.2018.2846735}
}

@article{Fujiwara2025,
  author       = {Takanori Fujiwara and
                  Kostiantyn Kucher and
                  Junpeng Wang and
                  Rafael Messias Martins and
                  Andreas Kerren and
                  Anders Ynnerman},
  title        = {Adversarial attacks on machine learning-aided visualizations},
  journal      = {J. Vis.},
  volume       = {28},
  number       = {1},
  pages        = {133--151},
  year         = {2025},
  _url         = {https://doi.org/10.1007/s12650-024-01029-2},
  doi          = {10.1007/S12650-024-01029-2}
}

@article{Espadoto2019,
  author       = {Mateus Espadoto and
                  Rafael Messias Martins and
                  Andreas Kerren and
                  Nina S. T. Hirata and
                  Alexandru C. Telea},
  title        = {Toward a Quantitative Survey of Dimension Reduction Techniques},
  journal      = {{IEEE} Trans. Vis. Comput. Graph.},
  volume       = {27},
  number       = {3},
  pages        = {2153--2173},
  year         = {2019},
  _url         = {https://doi.org/10.1109/TVCG.2019.2944182},
  doi          = {10.1109/TVCG.2019.2944182}
}

@article{Cunningham2015,
  author       = {John P. Cunningham and
                  Zoubin Ghahramani},
  title        = {Linear Dimensionality Reduction: Survey, Insights, and Generalizations},
  journal      = {J. Mach. Learn. Res.},
  volume       = {16},
  pages        = {2859--2900},
  year         = {2015},
  _url         = {https://dl.acm.org/doi/10.5555/2789272.2912091},
  doi          = {10.5555/2789272.2912091}
}

@article{Shusen2016,  
  author       = {Shusen Liu and
                  Dan Maljovec and
                  Bei Wang and
                  Peer{-}Timo Bremer and
                  Valerio Pascucci},
  title        = {Visualizing High-Dimensional Data: Advances in the Past Decade},
  journal      = {{IEEE} Trans. Vis. Comput. Graph.},
  volume       = {23},
  number       = {3},
  pages        = {1249--1268},
  year         = {2016},
  _url         = {https://doi.org/10.1109/TVCG.2016.2640960},
  doi          = {10.1109/TVCG.2016.2640960}
}

@inproceedings{Madry2018,
  author       = {Aleksander Madry and
                  Aleksandar Makelov and
                  Ludwig Schmidt and
                  Dimitris Tsipras and
                  Adrian Vladu},
  title        = {Towards Deep Learning Models Resistant to Adversarial Attacks},
  booktitle    = {6th Int. Conf. Learn. Represent.},
  _publisher   = {OpenReview.net},
  year         = {2018},
  _url         = {https://openreview.net/forum?id=rJzIBfZAb}
}

@inproceedings{Goodfellow2015Explaining,
  author       = {Ian J. Goodfellow and
                  Jonathon Shlens and
                  Christian Szegedy},
  _editor      = {Yoshua Bengio and
                  Yann LeCun},
  title        = {Explaining and Harnessing Adversarial Examples},
  booktitle    = {3rd Int. Conf. Learn. Represent.},
  year         = {2015},
  url          = {http://arxiv.org/abs/1412.6572}
}

@inproceedings{Wong2020Fast,
  author    = {Wong, Eric and Rice, Leslie and Kolter, J. Zico},
  title     = {Fast is better than free: Revisiting adversarial training},
  booktitle = {8th Int. Conf. Learn. Represent.},
  year      = {2020},
  url       = {https://openreview.net/forum?id=BJx040EFvH}
}

@article{Lin2024,
  author    = {Lin, Justin and Fukuyama, Julia},
  title     = {Calibrating dimension reduction hyperparameters in the presence of noise},
  journal   = {PLoS Comput. Biol.},
  volume    = {20},
  number    = {9},
  pages     = {e1012427},
  year      = {2024},
  doi       = {10.1371/journal.pcbi.1012427},
  _pmid     = {39264943}
}

@article{Kabaha2024,
  author    = {Kabaha, Anan and Drachsler-Cohen, Dana},
  title     = {Verification of Neural Networks' Global Robustness},
  journal   = {{ACM} Program. Lang.},
  volume    = {8},
  number   = {OOPSLA1},
  articleno = {130},
  pages     = {1010--1039},
  year      = {2024},
  doi       = {10.1145/3649847}
}

@article{Ellis2007ClutterReduction,
  author  = {Ellis, Geoffrey and Dix, Alan},
  title   = {A Taxonomy of Clutter Reduction for Information Visualisation},
  journal = {{IEEE} Trans. Vis. Comput. Graph.},
  volume  = {13},
  number  = {6},
  pages   = {1216--1223},
  year    = {2007},
  doi     = {10.1109/TVCG.2007.70535}
}

@inproceedings{vonLandesberger2009TimeDependentPointClouds,
  author    = {von Landesberger, Tatiana and Bremm, Sebastian and Rezaei, Mohammad and Schreck, Tobias},
  title     = {Visual Analytics of Time Dependent {2D} Point Clouds},
  booktitle = {Comput. Graph. Int.},
  year      = {2009},
  pages     = {97--101},
  _url      = {https://doi.org/10.1145/1629739.1629751},
  doi       = {10.1145/1629739.1629751},
}

@article{Chen2014ScatterplotSamplingDesignSpace,
  author       = {Haidong Chen and
                  Wei Chen and
                  Honghui Mei and
                  Zhiqi Liu and
                  Kun Zhou and
                  Weifeng Chen and
                  Wentao Gu and
                  Kwan{-}Liu Ma},
  title        = {Visual Abstraction and Exploration of Multi-class Scatterplots},
  journal      = {{IEEE} Trans. Vis. Comput. Graph.},
  volume       = {20},
  number       = {12},
  pages        = {1683--1692},
  year         = {2014},
  _url         = {https://doi.org/10.1109/TVCG.2014.2346594},
  doi          = {10.1109/TVCG.2014.2346594}
}

@article{Liao2018ClusterBasedScatterplots,
  author       = {Hongsen Liao and
                  Yingcai Wu and
                  Li Chen and
                  Wei Chen},
  title        = {Cluster-Based Visual Abstraction for Multivariate Scatterplots},
  journal      = {{IEEE} Trans. Vis. Comput. Graph.},
  volume       = {24},
  number       = {9},
  pages        = {2531--2545},
  year         = {2018},
  _url         = {https://doi.org/10.1109/TVCG.2017.2754480},
  doi          = {10.1109/TVCG.2017.2754480}
}

@book{Silverman1986Density,
  author       = {Bernard W. Silverman},
  title        = {Density Estimation for Statistics and Data Analysis},
  publisher    = {Springer},
  year         = {1986},
  _url         = {https://doi.org/10.1007/978-1-4899-3324-9},
  doi          = {10.1007/978-1-4899-3324-9},
  isbn         = {978-1-4899-3324-9}
}

@book{Scott1992Multivariate,
  author       = {David W. Scott},
  title        = {Multivariate Density Estimation: {Theory}, Practice, and Visualization},
  _series       = {Wiley Series in Probability and Statistics},
  publisher    = {Wiley},
  year         = {1992},
  _url         = {https://doi.org/10.1002/9780470316849},
  doi          = {10.1002/9780470316849},
  isbn         = {978-0-47154770-9}
}

@inproceedings{Venna2001,
  author       = {Jarkko Venna and
                  Samuel Kaski},
  _editor       = {Georg Dorffner and
                  Horst Bischof and
                  Kurt Hornik},
  title        = {Neighborhood Preservation in Nonlinear Projection Methods: {An} Experimental
                  Study},
  booktitle    = {30th Int. Conf. Artif. Neural Netw.},
  _series      = {Lecture Notes in Computer Science},
  volume       = {2130},
  pages        = {485--491},
  _publisher   = {Springer},
  year         = {2001},
  _url         = {https://doi.org/10.1007/3-540-44668-0\_68},
  doi          = {10.1007/3-540-44668-0\_68}
}

@article{Ngo2022Machine,
  author     = {Ngo, Quynh Quang and Dennig, Frederik L. and Keim, Daniel A. and Sedlmair, Michael},
  title      = {{Machine learning meets visualization - Experiences and lessons learned}},
  journal    = {it - Information Technology},
  volume     = {64},
  number     = {4--5},
  pages      = {169--180},
  _publisher = {De Gruyter Brill},
  year       = {2022},
  doi        = {10.1515/ITIT-2022-0034},
  _url       = {https://www.degruyterbrill.com/document/doi/10.1515/itit-2022-0034/html}
}

@inproceedings{Kingma2015,
  author       = {Diederik P. Kingma and
                  Jimmy Ba},
  _editor      = {Yoshua Bengio and
                  Yann LeCun},
  title        = {Adam: {A} Method for Stochastic Optimization},
  booktitle    = {3rd Int. Conf. Learn. Represent.},
  year         = {2015},
  _url         = {http://arxiv.org/abs/1412.6980}
}

@inproceedings{Colange2020,
  author       = {Beno{\^{\i}}t Colange and
                  Jaakko Peltonen and
                  Micha{\"{e}}l Aupetit and
                  Denys Dutykh and
                  Sylvain Lespinats},
  _editor      = {Hugo Larochelle and
                  Marc'Aurelio Ranzato and
                  Raia Hadsell and
                  Maria{-}Florina Balcan and
                  Hsuan{-}Tien Lin},
  title        = {Steering Distortions to Preserve Classes and Neighbors in Supervised
                  Dimensionality Reduction},
  booktitle    = {Adv. Neural Inf. Process. Syst.},
  year         = {2020},
  pages        = {13214--13225},
  volume       = {33},
  _publisher   = {Curran Associates, Inc.},
  _url         = {https://proceedings.neurips.cc/paper/2020/hash/99607461cdb9c26e2bd5f31b12dcf27a-Abstract.html}
}

@inproceedings{Miyato2018Spectral,
  author       = {Takeru Miyato and
                  Toshiki Kataoka and
                  Masanori Koyama and
                  Yuichi Yoshida},
  title        = {Spectral Normalization for Generative Adversarial Networks},
  booktitle    = {6th Int. Conf. Learn. Represent.},
  _publisher   = {OpenReview.net},
  year         = {2018},
  _url         = {https://openreview.net/forum?id=B1QRgziT-}
}

@inproceedings{Jakubovitz2018,
  author       = {Daniel Jakubovitz and
                  Raja Giryes},
  _editor      = {Vittorio Ferrari and
                  Martial Hebert and
                  Cristian Sminchisescu and
                  Yair Weiss},
  title        = {Improving {DNN} Robustness to Adversarial Attacks Using Jacobian Regularization},
  booktitle    = {Comput. Vis.},
  _series      = {Lecture Notes in Computer Science},
  volume       = {11216},
  pages        = {525--541},
  _publisher   = {Springer},
  year         = {2018},
  _url         = {https://doi.org/10.1007/978-3-030-01258-8\_32},
  doi          = {10.1007/978-3-030-01258-8\_32}
}

@article{Aurenhammer1991,
  author       = {Franz Aurenhammer},
  title        = {Voronoi Diagrams - {A} Survey of a Fundamental Geometric Data Structure},
  journal      = {{ACM} Comput. Surv.},
  volume       = {23},
  number       = {3},
  pages        = {345--405},
  year         = {1991},
  _url         = {https://doi.org/10.1145/116873.116880},
  doi          = {10.1145/116873.116880}
}

@article{Espadoto2021Unprojection,
  author       = {Mateus Espadoto and
                  Gabriel Appleby and
                  Ashley Suh and
                  Dylan Cashman and
                  Mingwei Li and
                  Carlos Scheidegger and
                  Erik W. Anderson and
                  Remco Chang andd
                  Alexandru C. Telea},
  title        = {{UnProjection}: Leveraging Inverse-Projections for Visual Analytics
                  of High-Dimensional Data},
  journal      = {IEEE Trans. Vis. Comput. Graph.},
  volume       = {29},
  number       = {2},
  pages        = {1559--1572},
  year         = {2021},
  _url         = {https://doi.org/10.1109/TVCG.2021.3125576},
  doi          = {10.1109/TVCG.2021.3125576}
}

@inproceedings{Cohen2019Certified,
  author       = {Jeremy Cohen and
                  Elan Rosenfeld and
                  J. Zico Kolter},
  _editor      = {Kamalika Chaudhuri and
                  Ruslan Salakhutdinov},
  title        = {Certified Adversarial Robustness via Randomized Smoothing},
  booktitle    = {36th Int. Conf. Mach. Learn.},
  _series      = {Proceedings of Machine Learning Research},
  pages        = {1310--1320},
  _publisher   = {{PMLR}},
  year         = {2019},
  url          = {http://proceedings.mlr.press/v97/cohen19c.html}
}

% biblatex with biber
%\printbibliography                

\end{document}